\documentclass[10pt,twocolumn]{article} 
\usepackage{primeARXIV}
\usepackage{times}
\usepackage{graphicx}
\usepackage{amssymb}
\usepackage{multirow}
\usepackage{algpseudocode}
\usepackage{algorithm}
\usepackage{array}
\usepackage{float}
\usepackage{arydshln}
\usepackage{placeins}
\usepackage{url,hyperref}
\usepackage{amsmath}
\usepackage{tikz}
\usepackage[normalem]{ulem}
\usepackage[format=plain,labelformat=simple,labelsep=period,font=small,compatibility=false]{caption}
\usepackage[font=footnotesize,skip=3pt,subrefformat=parens]{subcaption}

\usepackage{cleveref}
\crefname{section}{Sec.}{Secs.}
\Crefname{section}{Section}{Sections}
\crefname{figure}{Fig.}{Figs.}
\Crefname{figure}{Figure}{Figures}
\Crefname{table}{Table}{Tables}
\crefname{table}{Tab.}{Tabs.}

\newcommand{\mycomment}[1]{}
\newcommand{\todel}{\mycomment}
\newcommand{\change}{}
\newcommand{\changebox}{}

\useunder{\uline}{\ul}{}

\newcommand\copyrighttext{%
  \footnotesize \textcopyright 2025 IEEE. Personal use of this material is permitted.
  Permission from IEEE must be obtained for all other uses, in any current or future
  media, including reprinting/republishing this material for advertising or promotional
  purposes, creating new collective works, for resale or redistribution to servers or
  lists, or reuse of any copyrighted component of this work in other works.}
\newcommand\copyrightnotice{%
\begin{tikzpicture}[remember picture,overlay]
\node[anchor=south,yshift=10pt] at (current page.south) 
  {\fbox{\parbox{\dimexpr\textwidth-\fboxsep-\fboxrule\relax}{\copyrighttext}}};
\end{tikzpicture}%
}

\begin{document}

\title{Deep priors for satellite image restoration with accurate uncertainties
\thanks{This work was partly supported by CNES under project name DEEPREG, and ANITI under grant agreement ANR-19-PI3A-0004.}}

\author{Maud Biquard\\
ISAE-Supaero / CNES\\
31400 Toulouse, France\\
{\tt\small maud.biquard@isae-supaero.fr}
\and
Marie Chabert\\
IRIT/INP-ENSEEIHT\\
31000 Toulouse, France\\
{\tt\small marie.chabert@toulouse-inp.fr}
\and
Florence Genin, Christophe Latry\\
CNES\\
31400 Toulouse, France\\
{\tt\small firstname.lastname@cnes.fr}
\and
% Christophe Latry\\
% CNES\\
% 31400 Toulouse, France\\
% {\tt\small christophe.latry@cnes.fr}
% \And
Thomas Oberlin\\
ISAE-Supaero\\
31400 Toulouse, France\\
{\tt\small thomas.oberlin@isae-supaero.fr}
}

\maketitle
\copyrightnotice
\thispagestyle{empty}

\begin{abstract}

\vspace{-1em}
Satellite optical images, upon their on-ground receipt, offer a distorted view of the observed scene. Their restoration, including denoising, deblurring, and sometimes super-resolution, is required before their exploitation. 
Moreover, quantifying the uncertainties related to this restoration helps to reduce the risks of misinterpreting the image content.
Deep learning methods are now state-of-the-art for satellite image restoration. Among them, direct inversion methods train a specific network for each sensor, and generally provide a point estimation of the restored image without the associated uncertainties. 
Alternatively, deep regularization (DR) methods learn a deep prior on target images before plugging it, as the regularization term, into a model-based optimization scheme. This allows for restoring images from several sensors with a single network and possibly for estimating associated uncertainties. 
In this paper, we introduce VBLE-xz, a DR method that solves the inverse problem in the latent space of a variational compressive autoencoder (CAE). We adapt the regularization strength by modulating the bitrate of the trained CAE with a training-free approach. Then, VBLE-xz estimates relevant uncertainties jointly in the latent and in the image spaces by sampling an explicit posterior estimated within variational inference. This enables fast posterior sampling, unlike state-of-the-art DR methods that use Markov chains or diffusion-based approaches.
We conduct a comprehensive set of experiments on very high-resolution simulated and real Pléiades images, asserting the performance, robustness and scalability of the proposed method. 
They demonstrate that VBLE-xz represents a compelling alternative to direct inversion methods when uncertainty quantification is required. The code associated to this paper is available in \url{https://github.com/MaudBqrd/VBLExz}.
\end{abstract}


\section{Introduction}
\label{sec:intro}

Satellite imaging is of considerable interest for various remote sensing applications such as environmental monitoring, conservation programs, or urban planning. 
Some satellite optical systems can now acquire images at a very high resolution.
For instance, a PLEIADES-HR satellite captures images in which each pixel represents a $70\text{cm}\times 70$cm area on the ground and is coded in 12 bits. This allows for a detailed visualization of the Earth's surface and provides extremely valuable information for object detection \cite{Han2014,Chen2014}, change detection, or semantic segmentation \cite{Yang2018,Benediktsson2005}.
However, despite constant technological advances in sensor accuracy, satellite images provide an imperfect representation of the observed scene. In particular, the images contain noise and blur induced by the optical system, satellite movement, atmospheric perturbations, and onboard compression. 
As a consequence, the received image must be restored to get the most out of its on-ground exploitation.
The restoration aims at enhancing the image quality by removing the noise and blur mentioned above and may also be associated with super-resolution or inpainting that ease downstream data interpretation. 
All these processes must preserve the integrity of the information: no information should be removed and no false details added. 
Furthermore, the restoration process has to remain computationally viable to treat the important amount of data transmitted on the ground. Finally, the restoration process should ideally be sufficiently generic to process images from different sensors.

We restrict our study to panchromatic images, which offer the highest spatial resolution. They can be combined with multi-spectral images of lower resolution through pansharpening to obtain high-resolution multispectral images.
The acquisition of a panchromatic image by the satellite optical system can be modeled as 
\begin{align}
    y = \mathcal{D}_0(h_0*l) + w \label{eq:ip},
\end{align}
where $l$ represents the observed landscape, $y$ the acquired image, $h_0 = h_{atmo} * h_{move} * h_{ins}$ models the combined effects of the atmosphere, of the movement during integration time, and of the instrument. $\mathcal{D}_0$ denotes the sampling operator. Vector $w$ represents a white Poisson-Gaussian noise which can, for the considered problems, be approximated by a Gaussian noise of variance $\sigma_w^2(l)=\sigma_0^2 + K \mathcal{D}_0(h_0*l)$, where $\sigma_0$ and $K$ are noise parameters that are specific to a given optical system. 
Current Pléiades image restoration processes \cite{Latry2012} employ traditional methods such as NL-Bayes for denoising \cite{Lebrun2013} and inverse filtering for deconvolution  \cite{Latry2012}. 
However, these methods require \textit{ad hoc} operations to stick to the white Gaussian noise model favorable to NL-Bayes denoiser \cite{Makitalo2012}.
Moreover, since the advent of deep learning, they are not state-of-the-art for image restoration anymore.

Indeed, deep learning has led to a substantial breakthrough in image restoration \cite{Zhang2021,Zhang2018a}. 
A first category of methods, which we denote by direct inversion methods in the sequel, learns a neural network that maps the degraded image to the restored image in a supervised manner \cite{Zhang2018a,Wang2022,Wang2018}. 
However, as they learn a specific mapping for each forward model defined by \eqref{eq:ip}, a neural network must be learned for each sensor. 
A second category of methods, which we call deep regularization (DR) methods in the following, merely learns a regularization independently of the forward model \eqref{eq:ip} from a dataset of target images, that is the original landscapes sampled at a target resolution. While plug-and-play (PnP) approaches learn an implicit regularization within a denoiser \cite{Zhang2021,Venkatakrishnan2013,Hurault2022,Kadkhodaie2021}, latent optimization methods regularize explicitly the inverse problem, by looking for the problem solution in the latent space of a generative model \cite{Bora2017,Biquard2023,Prost2023}. 
DR methods typically compute the Maximum A Posteriori (MAP) estimate:
\begin{equation}
    \max_x p_{X|Y} (x|y) \Leftrightarrow \min_x - \log p_{Y|X}(y|x) - \log p_X(x) \label{eq:map_ip}
\end{equation}
where $x$ is the current estimation of the target image, $y$ is the measured image, $p_{Y|X}(y|x)$ is the data likelihood, $p_{X|Y} (x|y)$ the inverse problem posterior. The regularization $- \log p_X(x)$ promotes solutions of the inverse problem most compatible with the prior $p_X(x)$, which is learned beforehand from the dataset of target images. 

Last but not least, some DR methods do not only estimate the MAP, but enable to sample from the posterior distribution $p_{X|Y} (x|y)$. This remarkably enables to derive uncertainties, which are valuable for downstream applications. However, these methods are often computationally expensive as they mainly rely on Markov Chain Monte Carlo (MCMC). To overcome this, in Biquard et al. \cite{Biquard2023}, we have proposed another method for approximate posterior sampling, called Variational Bayes Latent Estimation (VBLE).  
By approximating the inverse problem posterior in the latent space of a variational compressive autoencoder (CAE),
VBLE allows for approximate posterior sampling with no significant increase in computational cost and seems well-suited for remote sensing image restoration. However, VBLE possesses two remaining drawbacks. First, it requires training CAEs at different bitrates to adapt the regularization strength to the inverse problem difficulty.
Second, the modeling of the posterior distribution in the latent space fails to fully account for all uncertainties, such as the representation error between the true solution and its projection onto the range of the generative model.

In this paper, we introduce VBLE-xz, a new method derived from VBLE that addresses those two limitations. Besides, 
while direct inversion methods are more and more studied in the context of remote sensing, and currently used in on-ground segments~\cite{Zhu2020,Chouteau2022}, 
DR methods have been rarely considered in the associated literature. We show here that VBLE-xz, and more generally DR methods, yield numerous advantages compared to them. Precisely, our contributions are the following:
\begin{itemize} 
    \item We propose an improved version of VBLE, called VBLE-xz. In particular, VBLE-xz models the posterior distribution of the inverse problem jointly in the latent and in image space, leading to an improved prediction of the uncertainties by a large margin.
    \item To restore inverse problems of diverse difficulties with the same network, we combine VBLE-xz with a multirate CAE. Following the procedure proposed in \cite{Arezki2024}, a single CAE is trained once, at a high bitrate. Then, this bitrate can be simply modified, during inference and without any finetuning. We adapt this framework to satellite image restoration and we show that it adjusts remarkably well to the inverse problem difficulty.
    \item We conduct a comprehensive set of experiments on high-resolution satellite images simulated in a realistic manner and on real satellite images.
    These experiments demonstrate two important points. First, we show that VBLE-xz yields state-of-the-art results in terms of restoration performance, uncertainty quantification, and computational load compared to other recent posterior sampling methods. 
    Second, we also compare the DR methods VBLE-xz and PG-DPIR \cite{Biquard2025} to direct inversion in terms of performance and robustness, showing that they remain competitive in terms of metrics, but with fewer artifacts and improved robustness to modeling errors. This demonstrates that
    deep regularization is a compelling alternative to direct inversion for real satellite processing chains.
\end{itemize}

\section{Related works}
\label{sec:relworks}

\subsection{Image restoration techniques}

Image restoration is a well-known inverse problem, which has classically been addressed using variational methods. The key point of these methods, which consist in minimizing a functional composed of a data fidelity term and a regularization term such as in \cref{eq:map_ip}, is to choose a proper regularization. Classical regularizations include total variation \cite{Rudin1992}, Tikhonov regularization \cite{Tikhonov1963}, and sparsity-promoting penalties on well-chosen representations such as wavelet bases or dictionaries \cite{Tibshirani1996,Elad2010,Mallat2008}. Recently, %Then, 
deep learning-based methods have emerged as a powerful alternative to these classical methods. Direct inversion methods, which directly learn a mapping from the degraded image to the restored image using a neural network, are widespread in image restoration. %imaging. 
They commonly use convolution neural networks (CNNs) within either a U-Net architecture \cite{Wang2022,Zamir2021,Luo2023}, or a ResNet architecture \cite{Zhang2017,Ledig2017,Wang2018}. 
Generative Adversarial Networks (GANs) have also been used to improve perceptual performance \cite{Ledig2017,Wang2018}. Recently, transformer-based methods have enabled better results \cite{Wang2022,Zamir2022,Liang2021}, as well as diffusion-based methods \cite{Lugmayr2022,Luo2023}, yet at the cost of an increased computational complexity. \change{Most of the above mentioned methods assume known degraded-target image pairs, where the degraded images, often simulated, may not be totally realistic. Therefore, a domain gap may exist when applying those methods on real-world problems. Blind methods, such as \cite{Lin2024,Xiao2025}, aim to jointly restore the image and estimate the degradation model. }

Unlike direct inversion methods, DR methods learn only the regularization term, with the help of denoisers or diffusion models in PnP methods \cite{Venkatakrishnan2013}, or with generative models such as Variational Autoencoders (VAEs) \cite{DiederikPKingma2014} or GANs \cite{Goodfellow2014} for latent optimization methods. PnP methods, which yield state-of-the-art results on a wide variety of inverse problems, consist in using splitting algorithms, such as Alternating Direction Method of Multipliers (ADMM) \cite{Venkatakrishnan2013,Chan2016} or Half Quadratic Splitting (HQS) \cite{Zhang2021,Hurault2022}, that separately handle the data term and the regularization term in the optimization problem. Their key concept is to use Gaussian denoisers \cite{Kamilov2017}, in particular deep denoisers \cite{Zhang2017a} in place of the proximal operator of the regularization. Recently, the use of denoising diffusion models \cite{Ho2020} has led to remarkable results for solving imaging inverse problems \cite{Kawar2022, Chung2022, Zhu2023}.

While PnP methods implicitly learn the regularization through the denoising process, latent optimization methods represent a more direct way to regularize inverse problems by estimating the data distribution within generative models. The seminal work \cite{Bora2017} seeks the inverse problem solution in the latent space of a VAE or of a GAN, by optimizing a functional, leading to a latent MAP estimate. Then, the solution is taken as the image generated from the latent MAP estimate. This approach is interesting because of its explicit formulation as well as its well-defined Bayesian framework. However, it suffers from its dependency on the quality of the generative model since the obtained solution necessarily lies in the generator range. To tackle this problem, \cite{Dhar2018,Dean2020} permit small deviations from the generative manifold while \cite{Gonzalez2022,Duff2022,Duff2023} propose to jointly optimize $x$ and $z$ using splitting algorithms. More recently, in \cite{Biquard2023}, we proposed to replace the generative model by a compressive autoencoder, which can generate a wide variety of images while efficiently regularizing the inverse problem in the latent space through its hyperprior.

While most of the methods described above only provide a single point estimate of the inverse problem, many applications, particularly in Earth observation, require an estimation of the confidence in this solution. 
To this end, several Bayesian methods enable to sample the posterior distribution $p_{X|Y}(x|y)$ of the inverse problem solution.
Some methods yield stochastic solutions to the inverse problem, relying, for instance, on the implicit prior provided by Gaussian denoisers \cite{Kadkhodaie2021} or exploiting diffusion model properties \cite{Zhu2023}. 
Another type of method uses Markov Chain Monte Carlo (MCMC) to sample from the true posterior. 
In particular, \cite{Durmus2018} proposes to use Unadjusted Langevin Algorithm (ULA) to solve imaging inverse problems.  Building on this, PnP-ULA \cite{Laumont2022a} approximates the log-likelihood gradient in ULA by employing a Gaussian denoiser in a PnP framework. Interestingly, \cite{Holden2022} attempts to combine MCMC with latent optimization methods by designing an MCMC sampling scheme in the latent space of a generative model.
These approaches enable to sample from the true posterior but require a lot of iterations to converge, making them hardly scalable for satellite imaging. 
\change{Alternatively, \cite{Marinescu2020,Dasgupta2024,Kothari2021,Biquard2023} propose to learn an approximate posterior distribution in the latent space of a generative model using variational inference. These approaches are usually much faster than other posterior sampling methods. In particular, in \cite{Biquard2023}}, we introduced the VBLE algorithm, performing variational inference in the latent space of a compressive autoencoder. With negligible additional computation cost compared to the deterministic algorithm, VBLE represents a relevant alternative to MCMCs for posterior sampling.

\subsection{Application to satellite optical imaging}
\label{sec:relworks_sat}

Once transmitted to the ground station and decompressed, satellite images are subject to an in-depth semantic analysis including for instance perpixel classification \cite{Yang2018,Benediktsson2005}, object detection \cite{Han2014,Chen2014}, or 3D reconstruction \cite{Facciolo2017}. Previously, satellite images must be restored and eventually super-resolved.
Most of the deep learning methods currently used are based on direct inversion. 
For satellite single image super-resolution (SISR) tasks, CNN-based methods \cite{Zhang2013,Liebel2016,Pouliot2018,Fu2018} and GAN-based methods \cite{Bosch2018,Jiang2019} are widely used.
In particular, \cite{Zhu2020,Chouteau2022} employ an SRResNet architecture for joint denoising, deblurring, and super-resolution of optical satellite images. 
Furthermore, \cite{Huang2015} proposes a deep pansharpening network, and \cite{Zhang2020,Zheng2021} perform deep hyperspectral super-resolution. 

Concerning DR methods, PnP methods are commonly used for hyperspectral image processing tasks. For instance, \cite{Wang2023,Lai2022} use ADMM PnP framework for hyperspectral image restoration, and \cite{Teodoro2017} for hyperspectral image sharpening. To process panchromatic or multispectral satellite images, \cite{Liu2022a} employs ADMM for pansharpening, while \cite{Tao2020} employs DPSR algorithm \cite{Zhang2019} for super-resolution. Up to our knowledge, PnP methods have not been commonly used for high-resolution satellite image restoration. Concerning latent optimization methods, they have been successfully applied to medical imaging \cite{Duff2023}. However, due to the complexity of using these methods on less structured datasets, they have yet not been applied to satellite imaging.

For satellite image restoration and super-resolution, the training data are crucial as the network has to generalize to real satellite data. Yet, getting realistic target images, or pairs of degraded and target images for training is challenging. Indeed, obtaining a proper target image requires downsampling an original acquired image by a sufficient factor to neglect the original acquisition effects. Thus, it is common~\cite{Liebel2016,Bosch2018,Jiang2019} to perform super-resolution at an unrealistic resolution, by considering the satellite image as the target and downsampling it with a bicubic kernel to get the low-resolution image. \cite{Pouliot2018} employ Sentinel-2 images at a resolution of 10m as high-resolution targets to super-resolve Landsat-5 images, which are at 30m resolution. In a similar way, \cite{SalgueiroRomero2020} uses Sentinel-2 images as input and Worldview images downsampled at a resolution of 2m as outputs, while \cite{Galar2019} learns to map Sentinel-2 images to RapidEye images at a resolution of 5m. For these approaches, the datasets are very limited, as the landscape and season should match for both satellites. \cite{Michel2022} presents an open-data licensed dataset composed of 10m and 20m surface reflectance patches from Sentinel-2, with their associated 5m patches acquired by the VEN$\mu$s satellite on the same day. 
Interestingly, \cite{Zhu2020,Chouteau2022} simulate realistic target and degraded images from aerial images that were originally at an extremely high resolution. They respectively use Google and Airbus owned aerial images, which are typically at a resolution lower than 10cm.

\subsection{Position to related works}

To our knowledge, this work is the first to evaluate DR methods on high-resolution satellite image restoration tasks. Most existing works in remote sensing rely on direct inversion approaches, as in \cite{Chouteau2022,Zhu2020}. However, DR methods offer several advantages over direct inversion for remote sensing image restoration. In particular, direct inversion typically requires training a dedicated network for each sensor, along with a computationally intensive data simulation process, as realistic pairs of degraded and reference images must be generated under a variety of acquisition conditions.
In contrast, DR methods enable a single neural network to be trained once and applied across multiple restoration tasks and sensors, aligning with the philosophy of foundation models~\cite{Bommasani2021}. Since this model is trained solely on target images, it significantly simplifies the data simulation pipeline.
Furthermore, our work involves a careful design of realistic degraded and reference satellite images, generated from very high-resolution airborne images. This stands in contrast to many existing studies in remote sensing, where unrealistic degradation models are often used due to the scarcity of reference data~\cite{Liebel2016,Bosch2018,Jiang2019}.

Then, VBLE-xz is a variational inference method. The key advantage of these methods for image restoration is that they enable very efficient posterior sampling by explicitly approximating the posterior. On the contrary, diffusion-based methods \cite{Kawar2022,Chung2022,Zhu2023} require to run a diffusion process for each sample to draw, while MCMC methods \cite{Holden2022,Laumont2022a} run a Markov chain requiring thousands of iterations to converge. This makes them computationally heavy and, in practice, hardly scalable for satellite image restoration pipelines.

Last, while variational inference has been commonly used for solving specific inverse problems \cite{Tonolini2020,Gao2022} in a supervised or semi-%(semi-)
supervised setting, 
\todel{VBLE \cite{Biquard2023} and VBLE-xz are the only variational inference methods leveraging latent optimization for solving several inverse problems with the same network. Yet VBLE-xz significantly improves VBLE method. First, by improving the uncertainty modeling, VBLE-xz improves both posterior samples and uncertainty quantification. Second, the use of a multi-rate CAE \cite{Arezki2024} to efficiently adapt the regularization strength enables to solve inverse problems of different difficulty with the same network, which was not the case for VBLE.}
\change{some approaches \cite{Kothari2021,Dasgupta2024,Marinescu2020,Biquard2023} propose to use it in a latent optimization setting. Precisely, in \cite{Kothari2021,Dasgupta2024}, small normalizing flows \cite{Dinh2015} are used to approximate the posterior in the latent space of a normalizing flow or a GAN. Alternatively, \cite{Marinescu2020,Biquard2023} are more closely related to VBLE-xz as they learn a simple unimodal approximate posterior in the latent space of a VAE or of a CAE. However, VBLE-xz significantly differs from these approaches. First, all the considered methods approximate the posterior distribution only in the latent space, while VBLE-xz approximates the joint latent and image posterior. 
Although \cite{Gonzalez2022, Prost2023} also leverages this joint posterior to compute a deterministic Maximum A Posteriori, VBLE-xz approximates this joint posterior approximation using variational inference, which is completely novel. Second, the use of a multi-rate CAE \cite{Arezki2024} to efficiently adapt the regularization strength enables to solve inverse problems of different difficulty with the same network. To our knowledge, such an approach to adapt the regularization has never been done for 
image restoration. }

\section{Variational Bayes joint latent-image estimation with accurate uncertainties}
\label{sec:vblexz}

\begin{figure*}[t]
    \centering
    \includegraphics[width = 1\textwidth]{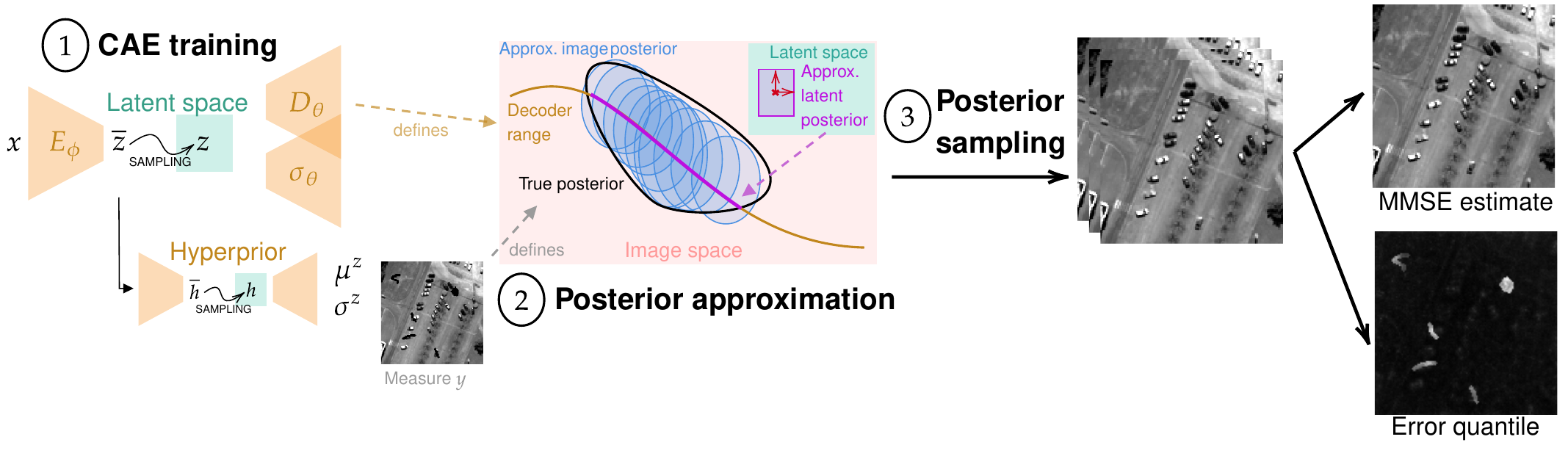}
    \caption{Principle of the proposed method, VBLE-xz. First, a CAE is trained on target images. Then, given a degraded image, the joint latent and image posterior is approximated following \cref{alg_ir}. Finally, posterior sampling is performed using the explicit approximate posterior, enabling to derive the posterior mean and uncertainty quantification maps.}
    \label{fig:abstract}
\end{figure*}

\subsection{Variational and compressive autoencoders}

\subsubsection{Variational autoencoders}
Variational autoencoders (VAEs) are generative neural networks, with a decoder $D_\theta$ learning a generative latent model and an encoder $E_\phi$ approximating the generative posterior distribution. Considering the following generative model:
\begin{equation}
    p_\theta (x,z) = p_\theta(x|z) p_\theta (z),
\end{equation}
$p_\theta (z)$ is a latent prior, often assumed to be $\mathcal{N}(0,I)$, while $p_\theta(x|z)$ represents the distribution learned by the decoder.  
The posterior distribution
\begin{equation}
    p_\theta(z|x) = \frac{p_\theta(x|z) p_\theta (z)}{\int_{z} p_\theta(x|z) p_\theta (z)}
\end{equation}
is often intractable, and is approximated by $q_\phi(z|x)$, the encoder distribution.

The weights $\theta$ and $\phi$ are learned during the training of the VAE by maximizing the Evidence Lower BOund (ELBO):

\begin{equation}
    \mathbb{E}_{q_\phi (z|x)} \big[ \log p_\theta(x|z) \big] - \text{KL}\big(q_\phi(z|x) || p_\theta(z)\big)
\end{equation}
where KL denotes the Kullblack-Leibler divergence between two distributions. In this paper, we consider $p_\theta(x|z) = \mathcal{N}(D_\theta(z), \Sigma_\theta(z))$, where $\Sigma_\theta(z)$ can be fixed, or learned. 
In classical VAEs, it is often assumed that $\Sigma_\theta(z) = \gamma^2 I$ and $q_\phi(z|x) = \mathcal{N}(\mu_\phi(x), \text{diag}(\sigma_\phi(x)^2))$. In this case, maximizing the ELBO amounts to minimizing the loss
\begin{align}
    \mathcal{L}(\theta,\phi,\gamma;x) = \mathbb{E}_{q_\phi (z|x)} \left[ \frac{1}{2\gamma^2} ||x - D_\theta(z)||^2_2 \right] \nonumber \\+ \text{KL}(q_\phi(z|x) || p_\theta(z|x)). \label{eq:vaeloss}
\end{align}
Hence, the VAE loss is composed of a data fidelity term and a KL divergence term, which constrains the distribution over the latent space. The parameter $\gamma$ tunes the trade-off between these two terms and can be either fixed or learned jointly.

\subsubsection{Variational compressive autoencoders}
Compressive autoencoders (CAEs) are neural networks used for compression, which have similarities to VAEs. CAEs are trained by optimizing the so-called rate-distortion trade-off. 
\begin{equation}
    \mathcal{L} = \mbox{ Rate}+\alpha \times \mbox{Distortion} \label{eq:cae_loss}.
\end{equation}
The distortion represents an error between the input and output of the CAE, and the rate measures the compression efficiency. CAEs can be expressed as VAEs with specific inference and generative distributions \cite{Balle2017}. In particular, the encoder posterior $q_\phi(z|x)$ is a uniform distribution, of mean $\bar z$ and with a fixed variance in order to simulate the quantization process.
Compared to VAE loss, the distortion resembles the data fidelity term, while the rate corresponds to the latent constraint. The trade-off parameter $\alpha$ plays a similar role as $\gamma$ for VAEs, controlling the trade-off between the two loss terms.

In addition, state-of-the-art CAEs use a hyperprior module \cite{Balle2017, Balle2018, Cheng2020}, as described in the first part of \cref{fig:abstract}, which is learned jointly with the main autoencoder. This additional autoencoder takes $\bar z$ as input and predicts $z$'s mean $\mu^z$ and deviation $\sigma^z$. The latent prior on $z$ is then defined as the factorized distribution $\prod_k \mathcal{N}(\mu^z_k,(\sigma^z_k)^2)$. Hence, the latent prior is a $z$-adaptive distribution, more powerful and flexible than the traditional VAE prior $\mathcal{N}(0,I)$. Formally, these CAEs can be expressed as VAEs with two latent variables $z$ and $h$ \cite{Balle2018}.

\subsubsection{Multi-rate CAEs}

To modulate the CAE compression rate, different CAEs should be trained at different values of $\alpha$ in \cref{eq:cae_loss}. However, several methods exist to design CAEs at multiple rates. The majority of these approaches \cite{Yang2020,Song2021,Arezki2024a,Mijares2023} propose to adapt the CAE structure and then to train the resulting network with a range of $\alpha$ values. Interestingly, \cite{Arezki2024} proposes a training-free adaptation of a CAE that was initially trained at a high bitrate. Indeed, given such a CAE, \cite{Arezki2024} introduces a scaling parameter $s \in ]0,1]$ at inference. Then, the auto-encoding process consists of:
\begin{align}
    z \sim q_\phi(z|s\times x), \hat x=D_\theta(z)/s. \label{eq:mr_cae}
\end{align}

\subsection{Background on VBLE with CAEs}

Variational Bayes Latent Estimation (VBLE) \cite{Biquard2023} is an algorithm used for image restoration, which enables efficient approximate posterior sampling. It belongs to latent optimization methods \cite{Bora2017}, which estimate the solution of an inverse problem in the latent space of a generative model. In place of a classical generative model, \cite{Biquard2023} proposed to use CAEs trained on image compression tasks. 

Furthermore, VBLE estimates the latent inverse problem posterior $p_{Z|Y}(z|y)$ in the latent space of the CAE using variational inference. Specifically, it considers the following factorized parametric family:
\begin{align}
    E_{\bar z,a} &= \left\{ q_{\bar z,a}(z) \big| \bar z,a \in \mathbb{R}^{C\times M\times N}, a>0 \right\}  \label{eq:paramfam} \\
    \text{ with } q_{\bar z,a}(z) &= \prod_k \mathcal{U}(z_k;[\bar z_k-\frac{a_k}{2},\bar z_k+\frac{a_k}{2}]) \nonumber
\end{align}
with $C$, $M$, and $N$ being the channel and spatial dimensions of the latent space.
The parametric distribution $q_{\bar z,a}$ is chosen to be uniform as it imitates the CAE encoder posterior $q_\phi(z|x)$.
Parameters $(\bar z, a)$ are classically optimized using variational inference by minimizing $\text{KL}(q_{\bar z,a} || p_{Z|Y})$.
It amounts to maximizing the corresponding ELBO:
\begin{align}
    \mathcal{L}(\bar z,a;y) = \mathbb{E}_{q_{\bar z,a} (z)} \big[ \log p_{Y|Z}(y|z) 
    + \lambda(\log p_\theta (z) \nonumber \\ - \log q_{\bar z,a}(z)) \big]  \label{eq:vble}
\end{align}
where $\lambda$ is a hyperparameter close to 1 which is added to adapt the regularization strength, $p_\theta (z)$ is the latent prior, that is the hyperprior $\mathcal{N}(\mu^z,\sigma^z)$ in the case of CAEs, with the notation of \cref{fig:abstract}.
Using the reparameterization trick $z = \bar z + a u$ with $u\sim \mathcal{U}(-\frac{1}{2},\frac{1}{2})$ \cite{DiederikPKingma2014}, \cref{eq:vble} can be optimized with a Stochastic Gradient Variational Bayes (SGVB) algorithm~\cite{DiederikP.Kingma2018}.

\subsection{Limitation of latent posterior estimation}

\begin{figure}[h]
            \centering
            \includegraphics[width = 0.4\textwidth]{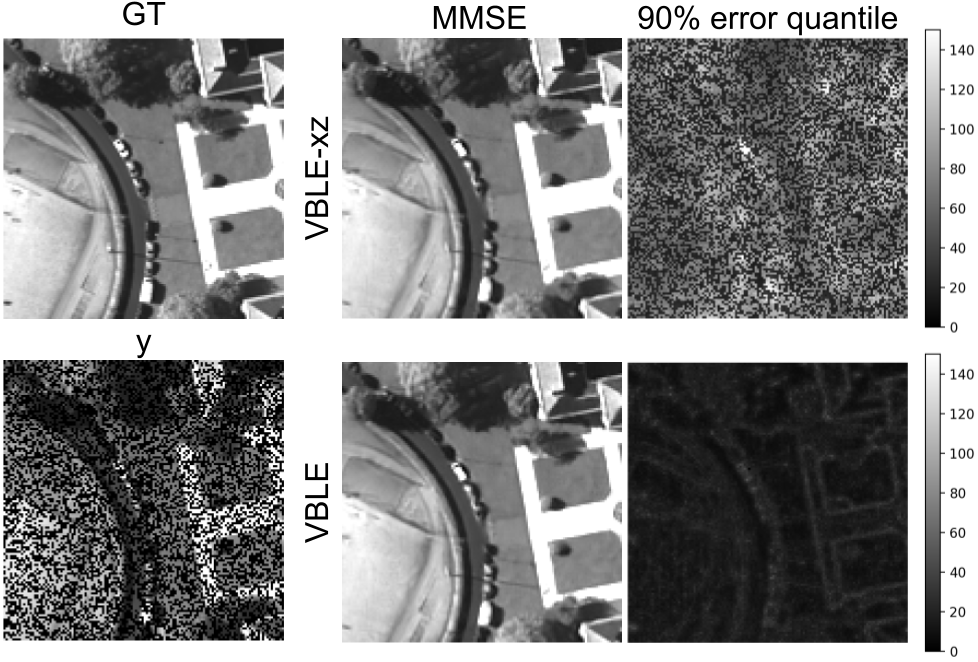}
            \caption{Visual result of VBLE and VBLE-xz on an inpainting problem. MMSE is the posterior mean computed on 100 posterior samples. The error quantile is a perpixel predicted error quantile computed with 100 posterior samples. \copyright CNES 2025}
            \label{fig:vble_vs_vblexz_uncertainty}
\end{figure}

In practice, the $\lambda$ parameter is not enough to adapt the regularization for inverse problems with very different levels of ill-posedness. A much more efficient way to control the regularization strength is the CAE bitrate, as a strong compression leads to a significant regularization. Hence, a first limitation of VBLE is the requirement to train CAEs at several bitrates to tackle inverse problems of various difficulties.

Furthermore, estimating the inverse problem uncertainty only in the latent space of the CAE works well in general but suffers from two main limitations. First, this prevents from modeling the representation error, that occurs when the solution does not lie in the decoder range. Second, the latent space posterior fails at capturing high-frequency patterns.
This specific point is illustrated in \cref{fig:vble_vs_vblexz_uncertainty}: for an inpainting problem with a random mask, VBLE does not recover the mask in its predicted error quantile. In contrast, VBLE-xz, which we introduce in the next section, enables to identify the mask, assigning greater uncertainty to pixels that have not been observed.

\subsection{VBLE-xz: Proposed joint latent and image posterior model}

\subsubsection{VBLE-xz algorithm}
In the following, we propose a new method for image restoration, derived from VBLE, in which we model the inverse problem posterior jointly in the latent and in the image spaces. 

We suppose that a CAE was trained on target images. We also assume that the generative model of this CAE is $p_\theta(x|z) = \mathcal{N}(D_\theta(z), \Sigma_\theta(z) := \operatorname{diag}(\sigma_\theta(z)^2)$, as illustrated in \cref{fig:abstract}. The training of the decoder that produces $\sigma_\theta(z)$ will be detailed later.
For simplicity, we only consider one latent variable $z$, but the equation can easily be adapted for two latent variable CAEs by considering a concatenated variable $\psi = (z,h)$ and $p(\psi) = p(z|h)p(h)$. 
\begin{figure}[t]
    \centering
    \includegraphics[width = 0.45\textwidth]{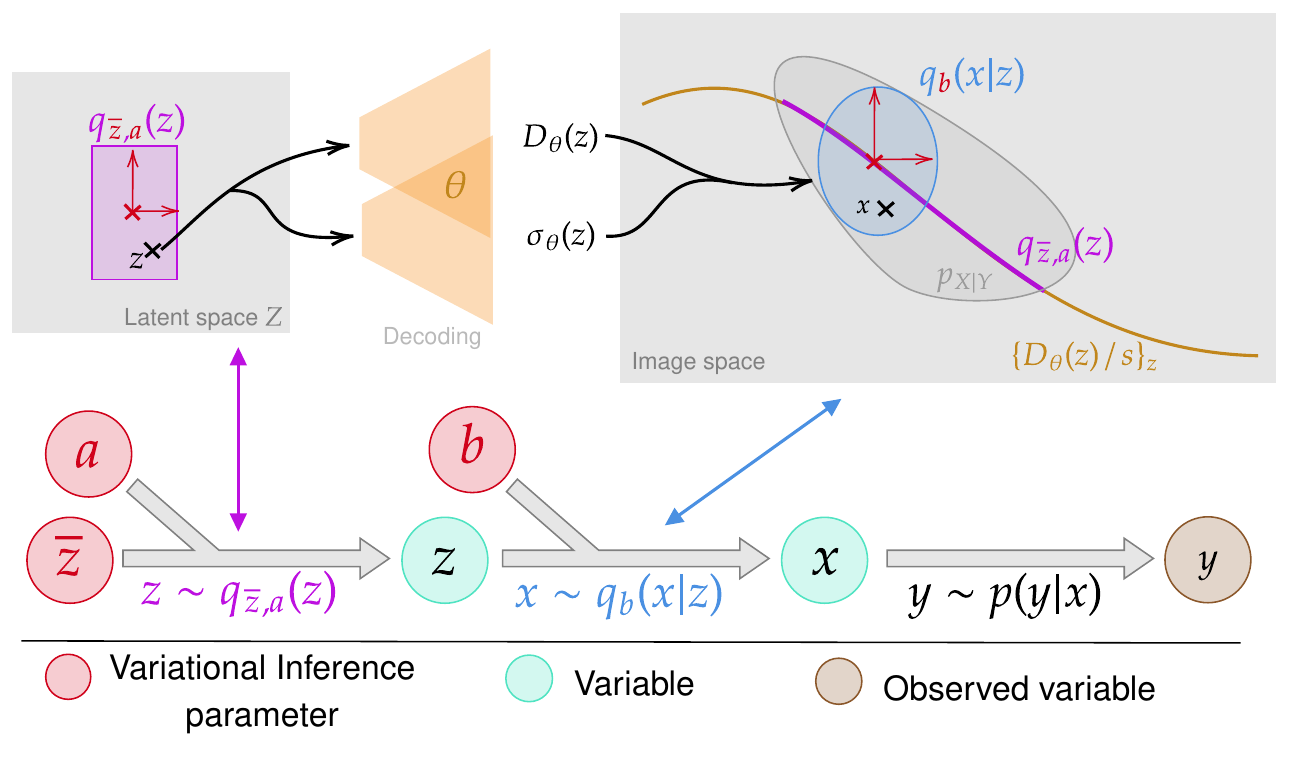}
    \caption{Generative model graph of our variational inference algorithm. The latent sampling from $q_{\bar z,a}$ and the image sampling from $q_b(x|z)=\mathcal{N}(x;D_\theta(z)/s,b\sigma_\theta(z)/s)$ are illustrated in the upper part of the Figure.}
    \label{fig:vblexz}
\end{figure}

We aim to estimate the joint latent and image inverse problem posterior $p_{X,Z|Y}(x,z|y)$ using variational inference. Note that, for VBLE, the targeted distribution was the latent posterior $p_{Z|Y}(z|y)$. For that purpose, we consider the following parametric family, corresponding to the representation of \cref{fig:vblexz}:
\begin{align*}
    E_{\bar z, a, b} &= \{ q_b(x|z) q_{\bar z, a}(z) | \bar z,a \in \mathbb{R}^{C\times M\times N},  \\
    & b \in  \mathbb{R}^{C'\times M'\times N'} a,b>0\} \\
    \text{with } & q_{\bar z,a}(z) = \prod_k \mathcal{U}(z_k;[\bar z_k-\frac{a_k}{2},\bar z_k+\frac{a_k}{2}]).  \nonumber \\
    & q_b(x|z) =  \prod_k \mathcal{N}\left(D_\theta(z)_k, b_k^2 \sigma_\theta(z)_k^2 \right)
\end{align*}
with $C',M'$ and $N'$  being the channel and spatial dimensions of the image space. The parametric distribution is composed of a latent term $q_{\bar z,a}(z)$ and of an image term $q_b(x|z)$ which was designed to take into account the representation error. As in VBLE, the latent term is chosen to be uniform as it matches the CAE encoder distribution. The choice for $q_b(x|z)$ is motivated by the fact that $p_\theta(x|z)$ represents the likelihood of generating $x$ given $z$. Hence, even if the representation error for an image $x^*$ is nonzero (that is $D_\theta(z) \neq x^*$), $x^*$ should nevertheless lie in a zone where $p_\theta(x|z)$ is high. Then, parameter $b$ allows to better adapt the posterior model to the given inverse problem. This idea is visually illustrated in \cref{fig:vblexz}.

Then, variational inference parameters $(\bar z, a, b)$ are optimized by minimizing $\operatorname{KL}(q_b(x|z) q_{\bar z, a}(z)||p_{X,Z|Y}(x,z|y))$, which amounts to maximizing the ELBO:
\begin{align}
    &\mathcal{L}(\bar z,a,b;y) = \mathbb{E}_{q_b(x|z)q_{\bar z,a} (z)} \big[ \log p_{Y|X}(y|x) \nonumber \\
    &+ \lambda (\log p_\theta (x|z) + \log p_\theta (z) - \log q_b(x|z) - \log q_{\bar z,a}(z)) \big] \label{eq:vblexz}
\end{align}
with $p_\theta (z)$ being the latent prior of the generative model, that is the hyperprior in the CAE case, and $\lambda$ a hyperparameter close to one, added to control the regularization as in VBLE.
Then, the $(x|z)$ terms represent a KL divergence between two Gaussian distributions and can be easily simplified, as well as $\log q_{\bar z,a}(z)$ which is independent of $z$. Thus:
\begin{align}
    & \mathcal{L}(\bar z,a,b;y) = \mathbb{E}_{q_b(x|z)q_{\bar z,a} (z)} \big[ \log p_{Y|X}(y|x) \nonumber \\
    &+ \lambda \log p_\theta (z) \big] + \lambda \left( \sum_k \log a_k + \sum_i \log b_i - \frac{1}{2}b_i^2 \right). \label{eq:vblexz3}
\end{align}
With the reparameterization trick
\begin{align}
    z = \bar z + a u, u\sim \mathcal{U}\left(\frac{-1}{2},\frac{1}{2}\right) \nonumber \\
    x = D_\theta(z) + b\sigma_\theta(z) \epsilon, \epsilon \sim \mathcal{N}(0,I), \label{eq:trick}
\end{align}
\cref{eq:vblexz3} can be optimized using gradient descent with a Monte Carlo approximation of the ELBO gradient as in \cite{DiederikPKingma2014,Biquard2023}. 

\begin{algorithm}
\small{
\caption{Posterior approximation with VBLE-xz}\label{alg_ir}
\begin{algorithmic}
    \Require $y$ measure, ($\lambda,s$) hyperparameters, $\bar z_0\in \mathbb{R}^{C\times M\times N}$, $a_0 \in \mathbb{R}^{C\times M\times N}=(1)_{i,j,l}$, $b_0\in \mathbb{R}^{C'\times M'\times N'}=(1)_{i,j,l}$  $l=0$, $\eta > 0$, $sg$ stop-gradient operation.
    \While{not \textit{convergence}}
        \State $z = \bar{z}_l + a_l u$ with $u \sim \prod_k\mathcal{U}(-0.5,0.5)$
        \State $x =D_\theta(z)/s + (b_lsg(\sigma_\theta(z))/s)\epsilon$ with $\epsilon \sim \mathcal{N}(0,I_{C'\times M'\times N'})$
        \State $\begin{pmatrix} \bar z_{l+1}\\ a_{l+1}\\b_{l+1} \end{pmatrix} = \begin{pmatrix} \bar z_{l}\\ a_{l} \\ b_l \end{pmatrix} + \eta \nabla_{\bar z,a,b} \big[ \log p_{Y|X}(y|x)
    + \lambda(\log p_\theta (z) + \sum_k \log a_{l,k} + \sum_{j} \log b_{l,j} - \frac{1}{2}b_{l,j}^2)\big]$
        \State $l = l + 1$
    \EndWhile \\
    \Return $\bar z_l,a_l,b_l$
\end{algorithmic}}
\end{algorithm}

\subsubsection{Modulating the CAE bitrate in VBLE-xz}

We introduce an additional scaling parameter in VBLE-xz in order to modulate the CAE bitrate in the same manner of \cref{eq:mr_cae} \cite{Arezki2024}. This scaling enables to modulate the regularization of the inverse problem without having to train several CAEs at several bitrates.
Let $s \in ]0,1]$ be a hyperparameter. Introducing the scaling into VBLE-xz amounts to change the distributions over $(x|z)$ in the following way:
\begin{align}
    p_\theta(x|z) = \prod_k\mathcal{N}(D_\theta(z)_k/s, \sigma_\theta(z)_k^2/s^2), \nonumber\\
    q_b(x|z) = \prod_k\mathcal{N}(D_\theta(z)_k/s, b_k^2\sigma_\theta(z)_k^2/s^2).
\end{align}
It induces slight changes in the sampling procedure described in \cref{eq:trick}. Apart from this, the loss in equations \cref{eq:vblexz} and \cref{eq:vblexz3} stays unchanged. The whole VBLE-xz algorithm is summarized in \cref{alg_ir}.

\subsubsection{Learning the generator variance}
\label{sec:genvar}
The \cref{alg_ir} requires the following generative model: $p_\theta(x|z) = \mathcal{N}(D_\theta(z), \text{diag}(\sigma_\theta(z)^2)$, thus the generator should estimate a pixelwise variance $\sigma_\theta(z)$, as illustrated in \cref{fig:abstract}.
However, in classical VAE and CAE models, the decoder distribution is a Gaussian distribution $p_\theta(x|z) = \mathcal{N}(D_\theta (z), \gamma^2I)$ with a fixed perpixel standard deviation $\gamma$ responsible for the trade-off between the data fidelity and the latent constraint terms in the loss. This distribution is not appropriate in our case as the reconstruction error should be larger on high frequency areas than on large flat ones. 

\looseness=-1
Some approaches learn more representative generator distributions \cite{Rybkin2021,Duff2023,Dorta2018}. 
Following \cite{Rybkin2021}, we add a second decoder predicting $\sigma_\theta(z)$, which has the same structure as the original decoder $D_\theta$. We adopt a two stage training strategy as is typically done for training stability: first, the CAE is classically trained with a given $\gamma$ trade-off parameter. Then, the second decoder is trained by minimizing the negative log-likelihood $\mathbb{E}_{z\sim q_\phi(z[x)]}[- \log p_\theta(x|z)]$, the other CAE parts being fixed. 

During \cref{alg_ir}, we introduce a stop-gradient operation on the deviation map $\sigma_\theta(z)$, that is $\sigma_\theta(z)$ is treated as a constant with respect to $z$ at each iteration. Without this operation, we observed that the deviation map abnormally converged toward zero, which prevented the $b_k$ coefficients from being correctly estimated. This behavior is likely due to the strong over-parameterization of the latent space in CAEs. Since addressing this issue during CAE training is not straightforward, the stop-gradient constitutes a relevant alternative that fully resolves the problem.

% \subsection{Calibration of VBLE predicted uncertainties}
% \label{sec:pock}

% Our method yields valuable and informative uncertainties, as it will be shown in the experiments. 
% However our variational posterior does not exactly match the true inverse problem posterior. To further improve the uncertainty prediction, we also consider the calibration strategy proposed in~\cite{Narnhofer2024}, which was shown to propose consistent error bounds for different posterior sampling methods.
% Specifically, this approach consists of learning, through a calibration subset, the quantiles of the true reconstruction error conditioned on an estimate of the posterior variance.  
% Afterward, for a given pixel, the $\alpha$-quantile of the true reconstruction error given the pixel posterior variance represents an error bound of level $\alpha$. We refer to \cite{Narnhofer2024} for further details.

\section{Experimental setting}
\label{sec:setup}

In the experiments, we consider the processing of one panchromatic spectral band, typically spanning wavelengths from 450nm to 850nm. We choose the operating point of Pléiades satellites with a resolution of 50cm, as it constitutes a reference for very high-resolution optical satellite imaging in Europe, for which we have a very precise model of the degradation.
In this section, we present the considered inverse problems. Then, data collection and processing are detailed. 
Finally, the experiment setup is provided, including the evaluation metrics, the considered baselines, and the training details for the proposed methods. 

\subsection{Forward models and considered inverse problems}

We consider three inverse problems: image restoration (IR), that is deblurring and denoising, joint image restoration and super-resolution (IR+SISR), and inpainting, with random or structured masks. The inpainting problem is noiseless.   
For IR and IR+SISR, the blurring process can hardly be exactly modelled, as the observed landscape, as well as the blur kernel, are continuous. Hence, we consider the forward model to be:
\begin{equation}
    y = \mathcal{D}(h* x_{target}) + w \label{eq:ip_approx}
\end{equation}
where $y$ is the measure, $x_{target}$ the target image at the target resolution, $h$ the blur kernel sampled at the target resolution, $\mathcal{D}$ corresponds to a downsampling operator, 
and $w$ is the approximated Poisson-Gaussian noise of variance $\sigma^2_w(x_{target}) = \sigma_0^2 +K(\mathcal{D}(h * x_{target}))$. 
For IR, $\mathcal{D} = I$ and for IR+SISR $\mathcal{D}$ is a decimation operator by a factor of 2. 

\subsection{Detailed data description}

\begin{table}[h]
\centering
\footnotesize
\setlength{\tabcolsep}{4pt}
\begin{tabular}{lcccccc}
\hline
 & \#imgs & \begin{tabular}[c]{@{}c@{}}Source\\ resolution\end{tabular} & \#bits & \begin{tabular}[c]{@{}c@{}}Covered\\ area ($km^2$)\end{tabular} & \begin{tabular}[c]{@{}c@{}}Size\\ ($\downarrow 50$cm)\end{tabular} & \begin{tabular}[c]{@{}c@{}}Size\\ ($\downarrow 25$cm)\end{tabular} \\ \hline
\textbf{Pélican} & 95 & 10cm & 12 & 18.45 & 128MB & 512MB \\
\textbf{PCRS} & 510 & 5cm & 12 & 537.18 & 4.3GB & 17.2GB \\ \hline
\end{tabular}
\vspace{0.5cm}
\setlength{\tabcolsep}{6pt}
\begin{tabular}{lccc}
    \hline
    DATASET & \textbf{\#imgs} & \textbf{$y$ size} & \textbf{resolution} \\ \hline
    TEST30 & 30 & $820 \times 820$ & 50cm \\
    TEST14\_PS512 & 14 & $512 \times 512$ & 50cm \\
    TEST14\_PS256 & 14 & $256 \times 256$ & 50cm \\
    TEST30\_REAL & 30 & $820 \times 820$ & 70cm \\ \hline
    \end{tabular}
\caption{Top: PCRS (IGN) and Pélican (CNES) databases characteristics. $\downarrow$50cm (resp. 25cm) corresponds to the information of the database downsampled at a resolution of 50cm (resp. 25cm). Bottom: Characteristics of the test databases used in the experiments. TEST30\_REAL contains real satellite images, the other images are simulated.}
\label{tab:data}
\end{table}

\subsubsection{Test and train datasets}
In this paper, we use simulated data to train VBLE-xz and the baselines. 
For the image restoration experiments, both simulated and real satellite data are used.
The simulated images are produced using two databases: PCRS, provided by Institut national de l'information géographique et forestière (IGN) \cite{IGNPCRS}, and Pélican, provided by CNES. Both datasets are composed of airplane images acquired at a very high resolution, that is 5cm for PCRS and 10cm for Pélican. Target and degraded satellite images are then obtained by downsampling them. The characteristics of each database are provided in the first part of \cref{tab:data}.
For the simulated test set, a subset of 30 images from the Pélican dataset is chosen, representing various landscapes such as desert, industrial, and urban areas. All PCRS images and the other Pélican images are used in the training and validation set. 
Some tests are performed on subsets of this test set because of the high computational cost of some baselines. The various test subsets are detailed in the second part of \cref{tab:data}. The tests on real Pléiades images are performed on 30 of them. Their characteristics are also given in \cref{tab:data}. 

\subsubsection{Image simulation process}

\looseness=-1
Target images are obtained by downsampling very high-resolution PCRS and Pélican images 
at the target resolution, that is 50cm for IR and for inpainting, and 25cm for IR+SISR, after anti-aliasing filtering. 
A sufficiently high downsampling factor makes the initial airplane degradation negligible, thus allowing us to consider them as ideal, that is free from noise and blur. 
Degraded images for IR and IR+SISR are obtained by first applying the blur kernel, which models the satellite point spread function (PSF), to Pélican and PCRS aerial images. Second, they are downsampled at the satellite resolution, that is 50cm in all simulated experiments, subjected to the simulated instrument noise, and finally quantized on 12 bits.
Note that degraded images obtained with this simulation pipeline do not strictly follow the forward model \cref{eq:ip_approx} that does not account for PSF subsampling and image quantization. This places us in a realistic context where the forward model is imperfectly known.

\subsubsection{Characteristics of the real images}
Experiments on real data are performed using Pléiades 12 bits images. These images are acquired at a resolution of 70cm (but commercialized at 50cm after an upsampling that enhances the image robustness to post-processing). Note that, unlike the simulated images, the real ones have been compressed at 2.86bits/pixel \cite{Latry2012} and decompressed. At such a high rate, the impact of compression-decompression can also be considered as an admissible practical deviation from the theoretical model.

\subsection{Experiment setup}

\subsubsection{Training settings}
\looseness=-1
The neural networks used in VBLE-xz are trained at each considered target resolution, that is 50cm and 25cm for the IR and IR+SISR experiments on simulated data, 70cm and 35cm for the IR and IR+SISR experiments on real data. 
We choose CAE architecture from \cite{Minnen2018}. It is a CAE with a hyperprior, with a simple convolutional encoder and decoder. We employ a pretrained model  from compressAI library \cite{Begaint2020} that has been pretrained at a high bitrate and we finetune it on satellite images at a given target resolution for 150k iterations. Afterward, we train the variance decoder as described in \cref{sec:genvar} using satellite images during 100k iterations.

\begin{table}[h]
    \footnotesize
    \centering
    \begin{tabular}{lccccc}
    \hline
     & \textbf{SRResNet} & \textbf{RDN} & \textbf{DRUNet} & \textbf{DiffUNet} & \textbf{CAE} \\
    \#params & 1.3M & 22M & 32M & 552M & 25M \\ \hline
    \end{tabular}
    \caption{Number of parameters of the different networks used in the experiments. DiffUnet corresponds to the diffusion model used in DiffPIR.}
    \label{tab:params}
\end{table}

\subsubsection{Baselines}
First, for uncertainty quantification evaluation, we compare VBLE-xz to three posterior sampling methods: PnP-ULA \cite{Laumont2022a}, a state-of-the-art MCMC posterior sampling method that employs a denoiser as prior, DiffPIR \cite{Zhu2023}, a state-of-the-art diffusion-based method and VBLE \cite{Biquard2023}, which is a natural baseline for VBLE-xz. For PnP-ULA, we choose the DRUNet denoiser \cite{Zhang2021} and for VBLE we choose the same CAE as for VBLE-xz. DRUNet and the diffusion model are finetuned on satellite images at the target resolution for about 100k iterations.
Then, to assess the performance of VBLE-xz without uncertainty quantification, we use VBLE and DiffPIR as baselines, as well as the following other baselines.  
Bay+IF \cite{Latry2012} is a classical approach that is currently being used in the Pléiades ground segment. 
It is composed of two steps: denoising is performed using NL-Bayes \cite{Lebrun2013}, then inverse filtering is used for deblurring. Super-resolution, when considered, is performed using bicubic interpolation. 
Then, we consider two direct inversion baselines, SRResNet and RDN. SRResNet \cite{Ledig2017} is a light convolutional neural network composed of a succession of residual blocks. This network has already been adapted for satellite image restoration and SISR~\cite{Zhu2020,Chouteau2022}. RDN \cite{Zhang2018a} is a deeper and wider convolutional network, which uses residual blocks with dense connections. For both networks, we adopt the loss proposed in \cite{Chouteau2022}: $\mathcal{L}(x,\hat x) = \alpha \text{SmoothL1}(x,\hat x) + \beta \operatorname{DISTS}(x,\hat{x})$ where SmoothL1 is a smooth version of the L1 distance, and DISTS \cite{Ding2020a} is a perceptual loss. SRResNet and RDN are trained from scratch during 200k iterations, corresponding to convergence. At last, we consider PG-DPIR \cite{Biquard2025}, an adaptation of the PnP method DPIR \cite{Zhang2021} for Poisson-Gaussian noise. The same denoiser as for PnP-ULA is used. Note that DiffPIR is designed for Gaussian noise, thus we use the same procedure as for PG-DPIR to adapt it to Poisson-Gaussian noise.
In \cref{tab:params}, the number of parameters of all the neural networks used in the experiments is provided.

\subsubsection{Restoration hyperparameters}
For several methods, restoration hyperparameters are tuned by grid search on a validation set of 3 Pélican images for each inverse problem. This procedure was applied to: the regularization parameter $\lambda$ and the scaling $s$ for VBLE-xz, the regularization parameter for PG-DPIR, the regularization parameter, the denoiser noise level and the chain step size for PnP-ULA and the regularization parameter for DiffPIR. For PnP-ULA, $10^5$ iterations are performed for each restoration. For DiffPIR, 100 diffusion steps are performed and $\zeta$ is fixed to 0.9.

\subsubsection{Evaluation metrics}
\label{sec:metrics}
To evaluate the image restoration performance when ground truth is available, three metrics have been used: the Peak Signal-to-Noise Ratio (PSNR), the Structural SIMilarity (SSIM) \cite{Wang2004}, and the Learned Perceptual Image Patch Similarity (LPIPS) \cite{Zhang2018}. The PSNR represents the accuracy between the restored and target images, while the SSIM and LPIPS are respectively classical and deep learning perceptual metrics.
For no reference image quality assessment, we employ the Blind Referenceless Image Spatial Quality Evaluator (BRISQUE) \cite{Mittal2012}. BRISQUE provides a score between 0 and 100, based on the deviation of the statistic of some local features with respect to natural images, the lower being the better.
Finally, to evaluate quantitatively the predicted uncertainty quantification of posterior sampling methods, we compute Interval Coverage Probability (ICP) of level $\alpha$. It represents the proportion of the perpixel true error that is contained in the perpixel predicted error quantiles of level $\alpha$. Thus, the closer to $\alpha$, the better. 

\section{Experimental results}
\label{sec:results}

\subsection{Comparison between posterior sampling methods on simulated data }
\label{sec:posterior_sampling_exp}

      \begin{table}[t]
        \centering
        \scriptsize
% \begin{tabular}{llccccc}
% \hline
%  & \textbf{} & \textbf{PnP-ULA} & \textbf{VBLE} & \textbf{DiffPIR} & \textbf{DiffPIR-100} & \textbf{VBLE-xz} \\ \hline
% \parbox[t]{2mm}{\multirow{5}{*}{\rotatebox[origin=c]{90}{\textbf{IR}}}} & PSNR $\uparrow$ & \textbf{48.63} & 48.40 & 45.54 & {\ul 48.47} & 48.29 \\
%  & SSIM $\uparrow$ & \textbf{0.9943} & {\ul 0.9938} & 0.9875 & 0.9937 & 0.9937 \\
%  & LPIPS $\downarrow$ & {\ul 0.0216} & \textbf{0.0214} & 0.0540 & 0.0260 & 0.0217 \\
%  & ICP $90\%$ & {\ul 0.92} & 0.79 & x & \textbf{0.90} & {\ul 0.92} \\
%  & Time & 2h10m & \textbf{17.7s} & {\ul 24.5s?} & 40min50s? & 27.3s \\ \hline
% \parbox[t]{2mm}{\multirow{5}{*}{\rotatebox[origin=c]{90}{\textbf{IR+SISR}}}} & PSNR $\uparrow$ & 35.86 & {\ul 37.45} & 37.02 & \textbf{38.19} & 37.31 \\
%  & SSIM $\uparrow$ & 0.9121 & {\ul 0.9501} & 0.9376 & \textbf{0.9528} & 0.9484 \\
%  & LPIPS $\downarrow$ & 0.2272 & 0.1773 & 0.1807 & {\ul 0.1703} & \textbf{0.1691} \\
%  & ICP $90\%$ & {\ul 0.94} & 0.66 & x & 0.75 & \textbf{0.88} \\
%  & Time & 2h10m & \textbf{19.5s} & 34.0s? & 56min40s? & {\ul 29.4s} \\ \hline
% \parbox[t]{2mm}{\multirow{5}{*}{\rotatebox[origin=c]{90}{\textbf{Inpainting}}}} & PSNR $\uparrow$ &  & 36.09 & 35.15 &  & 36.48 \\
%  & SSIM $\uparrow$ &  & 0.9546 & 0.9350 &  & 0.9562 \\
%  & LPIPS $\downarrow$ &  & 0.1058 & 0.1324 &  & 0.1106 \\
%  & ICP $90\%$ &  & 0.67 & x &  & 0.90 \\
%  & Time &  & 13.7s? & 25.5s? &  & 14.2s? \\ \hline
% \end{tabular}
\setlength{\tabcolsep}{4pt}
\begin{tabular}{llccccc}
\hline
 & \textbf{} & \textbf{PnP-ULA} & \textbf{VBLE} & \textbf{DiffPIR} & \textbf{DiffPIR-100} & \textbf{VBLE-xz} \\ \hline
\parbox[t]{2mm}{\multirow{5}{*}{\rotatebox[origin=c]{90}{\textbf{IR}}}} & PSNR $\uparrow$ & \textbf{48.63} & 48.40 & 45.62 & {\ul 48.53} & 48.29 \\
 & SSIM $\uparrow$ & \textbf{0.9943} & {\ul 0.9938} & 0.9878 & \underline{0.9938} & 0.9937 \\
 & LPIPS $\downarrow$ & {\ul 0.0216} & \textbf{0.0214} & 0.0537 & 0.0263 & 0.0217 \\
 & ICP $90\%$ & {\ul 0.92} & 0.79 & x & \textbf{0.90} & {\ul 0.92} \\
 & Time & 2h10m & \textbf{23.5s} & x & 1h55min & \underline{29.5s} \\ \hline
\parbox[t]{2mm}{\multirow{5}{*}{\rotatebox[origin=c]{90}{\textbf{IR+SISR}}}} & PSNR $\uparrow$ & 35.86 & {\ul 37.45} & 36.99 & \textbf{38.29} & 37.31 \\
 & SSIM $\uparrow$ & 0.9121 & {\ul 0.9501} & 0.9374 & \textbf{0.9534} & 0.9484 \\
 & LPIPS $\downarrow$ & 0.2272 & 0.1773 & 0.1746 & 
 \textbf{0.1668}& \underline{0.1691} \\
 & ICP $90\%$ & {\ul 0.94} & 0.66 & x & 0.76 & \textbf{0.88} \\
 & Time & 2h10m & \textbf{25.6s} & x & 2h17min & {\ul 32.1s} \\ \hline
\parbox[t]{2mm}{\multirow{5}{*}{\rotatebox[origin=c]{90}{\textbf{Inpainting}}}} & PSNR $\uparrow$ & 36.45 & 36.09 & 35.25 & \textbf{36.79} & {\ul 36.48} \\
 & SSIM $\uparrow$ & \textbf{0.9597} & 0.9546 & 0.9361 & 0.9541 & {\ul 0.9562} \\
 & LPIPS $\downarrow$ & {\ul 0.1065} & \textbf{0.1058} & 0.1296 & 0.1198 & 0.1106 \\
 & ICP $90\%$ & 0.47 & 0.67 & x & \underline{0.79} & \textbf{0.90} \\
 & Time & 2h10min & \textbf{33.3s} & x & 1h57min & \underline{43.9s} \\ \hline
\end{tabular}
\caption{Comparison between posterior sampling methods on TEST14\_SIMU for IR, IR+SISR and inpainting. Time denotes the time required to restore and draw 100 posterior samples for each method given a $512^2$ image for IR and $256^2$ for IR+SISR. For VBLE-xz, it includes the restoration time of \cref{alg_ir} and the posterior sampling time. Times are obtained with a Nvidia Quadro RTX8000 GPU.}
        \label{tab:uncertainty}
        \vspace{-2em}
        \end{table}

        \begin{figure}[h]
    \begin{minipage}[b]{0.32\linewidth}
        \centering
        \includegraphics[width=\textwidth]{"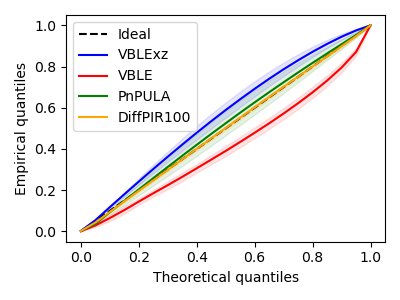"}      
       \end{minipage}
       \begin{minipage}[b]{0.32\linewidth}
        \centering
        \includegraphics[width=\textwidth]{"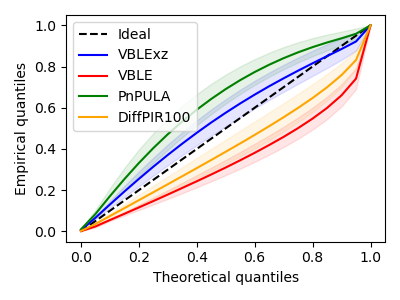"}      
       \end{minipage}\begin{minipage}[b]{0.32\linewidth}
        \centering
        \includegraphics[width=\textwidth]{"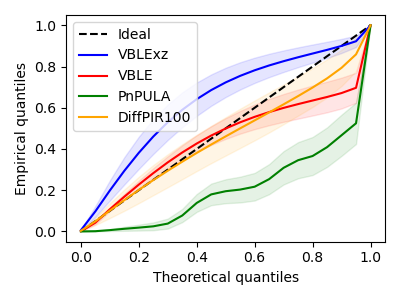"}     
       \end{minipage}

       \caption{Interval coverage probabilities for the IR problem (left), IR+SISR problem (middle) and inpainting (right). Each point of a curve of abscissa $\alpha$ provides the ICP of level $\alpha$ averaged on the test dataset. Colored areas correspond to deviations. The closer to the identity, the better.}
       \label{fig:coverage}
       \vspace{-1em}
\end{figure}

        % \begin{figure}[h]
        %     \centering
        %     \includegraphics[width = 0.5\textwidth]{figures/VBLE_vs_VBLExz_uncertainty.png}
        %     \caption{Difference between VBLE and VBLE-xz posterior sampling ability. %The true error is $e=|GT-MMSE|$ while 
        %     The error quantile map represents the perpixel 90\% quantile of the predicted error. \copyright CNES 2025}
        %     \label{fig:vble_vs_vblexz_uncertainty}
        % \end{figure}
    %     \begin{figure*}[h]
    %     \centering
    %     \includegraphics[width = 0.7\textwidth]{figures/new_sample1.png}
    %     \includegraphics[width = 0.7\textwidth]{figures/new_sample2.png}
    %     \caption{Posterior sampling ability of various methods on TEST14\_SIMU. Top: IR problem, Bottom: IR+SISR problem. MMSE estimates are averaged on 100 posterior samples. The true error is $e=|GT-MMSE|$ while the error quantile map represents the perpixel 90\% quantile of the predicted error. VBLE-xz quantiles are calibrated, as proposed in \cref{sec:pock}. \copyright CNES 2025}
    %     \label{fig:sampling1}
    % \end{figure*}
    \begin{figure*}[!htb]
    \centering
        \includegraphics[width = 0.85\textwidth]{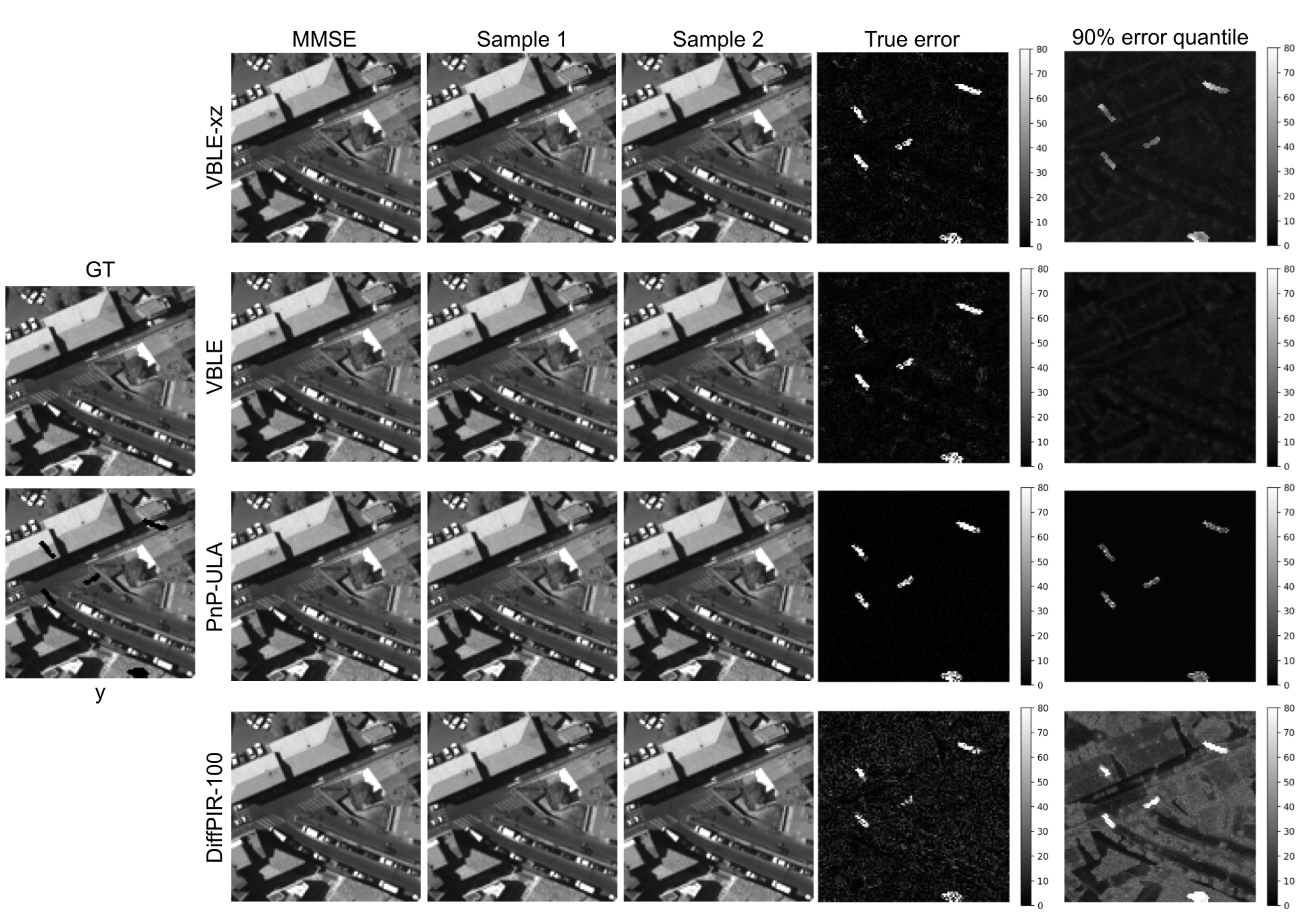}
        \includegraphics[width = 0.85\textwidth]{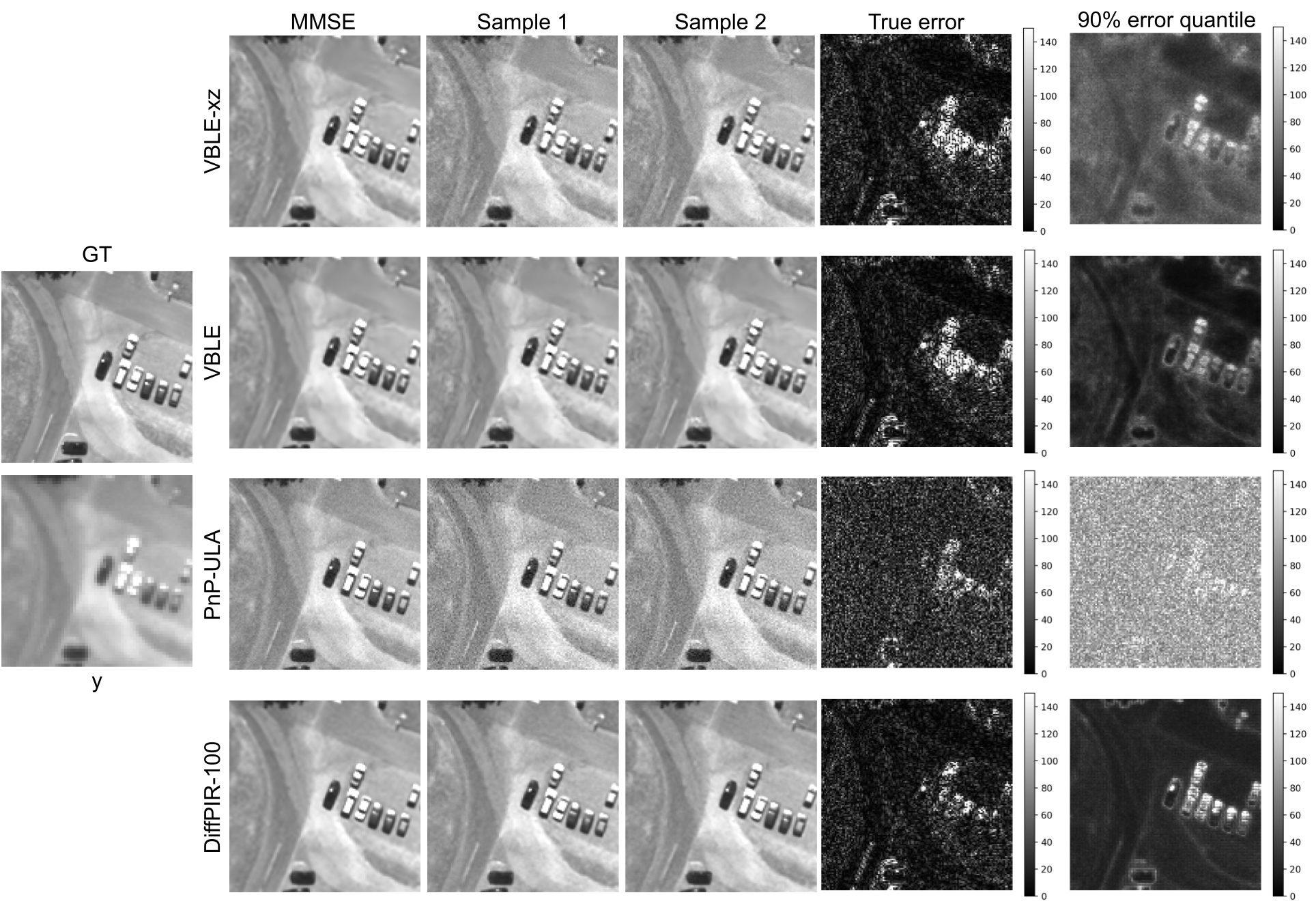}
        
    \caption{Posterior sampling ability of various methods on TEST14\_SIMU. Top: Structured inpainting problem. Bottom: IR+SISR problem. MMSE estimates are averaged on 100 posterior samples. The true error is $e=|GT-MMSE|$ while the error quantile map represents the perpixel 90\% quantile of the predicted error. \copyright CNES 2025}
    \label{fig:sampling1}
\end{figure*}

\looseness=-1
In this section, we assess the performance in restoration and for uncertainty quantification, for the posterior sampling methods. VBLE-xz is thus compared to DiffPIR \cite{Zhu2023, }VBLE \cite{Biquard2023} and PnP-ULA \cite{Laumont2022a}. 
We perform these experiments on TEST14 sets, for IR, IR+SISR and inpainting. 
We provide quantitative results in \cref{tab:uncertainty}, including the 90\% ICP (defined in \cref{sec:metrics}) as well as the computation time required to restore an image and draw 100 posterior samples. For DiffPIR, the results are given using one posterior sample, as classically done in the literature, and with the posterior
mean averaged on 100 samples, denoted as DiffPIR100.
First, a key advantage of VBLE-xz is the little time it takes to draw 100 posterior samples. 
Indeed, it is by several orders of magnitude faster than PnP-ULA and DiffPIR100, making this method scalable for satellite image restoration. 
In terms of PSNR and LPIPS, VBLE and VBLE-xz have very close results, with VBLE-xz being slightly better, especially in inpainting. This gain is not significant but is a positive outcome as VBLE-xz was primarily designed to enhance uncertainty quantification in VBLE. 
PnP-ULA exhibits consistent results but it is outperformed by VBLE-xz in super-resolution. DiffPIR is outperformed by VBLE-xz in all cases, while DiffPIR100 performs very well, however with a significant computation time, making it unrealistic for real satellite processing chains. 
In \cref{fig:coverage}, coverage probabilities, that is ICP of level $\alpha$ for $\alpha$ varying from 0 to 1, are provided for each method on the three problems. The 90\% ICP of \cref{tab:uncertainty} corresponds to the point with abscissa 0.9. 
First, VBLE-xz yields significantly better uncertainties than VBLE, validating the proposed joint latent and image posterior modeling. Besides, on average, VBLE-xz outperforms PnP-ULA and is close to DiffPIR100 in terms of ICP.

\looseness=-1
Lastly, a thorough visual study of each method's posterior sampling ability is provided in \cref{fig:sampling1}. This figure shows the MMSE estimate of each method obtained by averaging 100 posterior samples, as well as two posterior samples, the true error, and the 90\% error quantile map. 
The interest of VBLE-xz over VBLE is visually assessed in the inpainting problem, as the inpainting mask is not retrieved in VBLE predicted error. Indeed, this kind of uncertainty, which presents a very sharp structure, is hardly modeled in the latent space only. We note that VBLE-xz posterior samples exhibit a little bit of noise as some uncorrelated noise is added in the image space, but this does not affect the image interpretation.
In contrast, in the particular case of IR+SISR, PnP-ULA posterior samples are very noisy making them hardly usable in practice, while DiffPIR100 and VBLE-xz yield realistic posterior samples and predicted errors.

\looseness=-1
Hence, VBLE-xz provides better posterior samples and uncertainties than VBLE, validating the proposed joint latent and image posterior modeling. Furthermore, VBLE-xz is faster than DiffPIR100 and PnP-ULA by several orders of magnitude for posterior sampling. This makes it scalable to restore real satellite images, unlike those methods.   

\subsection{Point estimation results on simulated data}

\begin{table*}[t]
    \centering
    \scriptsize
    \setlength{\tabcolsep}{2pt}
    \begin{tabular}{l|cccccccc|ccccccc}
\hline
\multicolumn{1}{l|}{} & \multicolumn{7}{c|}{\textbf{IR}} & \multicolumn{6}{c}{\textbf{IR+SISR}} \\
 & \begin{tabular}[c]{@{}c@{}}Degraded\\ image $y$\end{tabular} & \textbf{Bay+IF} & \textbf{RDN} & \textbf{SRResNet} & \textbf{DiffPIR} & \textbf{VBLE} & \multicolumn{1}{c|}{\textbf{PG-DPIR}} & \textbf{VBLE-xz} & \textbf{\begin{tabular}[c]{@{}c@{}}Bay+IF\\ (+bicubic)\end{tabular}} & \textbf{RDN} & \textbf{SRResNet} & \textbf{DiffPIR} & \textbf{VBLE} & \multicolumn{1}{c|}{\textbf{PG-DPIR}} & \textbf{VBLE-xz} \\ \hline
PSNR $\uparrow$ & 33.55 & 40.56 & 48.08 & {\ul 48.44} & 45.48 & 48.38 & \multicolumn{1}{c|}{\textbf{48.66}} & 48.31 & 33.22 & {\ul 36.66} & 36.28 & 36.26 & 36.20 & \multicolumn{1}{c|}{\textbf{37.18}} & 36.65 \\
SSIM $\uparrow$ & 0.9289 & 0.9859 & {\ul 0.9951} & 0.9949 & 0.9893 & 0.9948 & \multicolumn{1}{c|}{\textbf{0.9952}} & 0.9948 & 0.9088 & 0.9460 & 0.9430 & 0.9345 & 0.9333 & \multicolumn{1}{c|}{\textbf{0.9513}} & {\ul 0.9486} \\
LPIPS $\downarrow$ & 0.1437 & 0.0369 & {\ul 0.0145} & 0.0157 & 0.0464 & 0.0207 & \multicolumn{1}{c|}{\textbf{0.0138}} & 0.0198 & 0.2463 & \textbf{0.1228} & {\ul 0.1466} & 0.1930 & 0.2018 & \multicolumn{1}{c|}{0.1658} & 0.1708 \\ \hline
\multicolumn{1}{l|}{Time} & x & x & 4.2s & \textbf{0.4s} & 3min34s & 49.8s & \multicolumn{1}{c|}{\underline{3.7s}} & 1min01s & x & \underline{5.0s} & \textbf{0.9s} & 18min42s & 4min05s & \multicolumn{1}{c|}{1min08s} & 5min10s \\ \hline
\end{tabular}
\caption{Quantitative results on TEST30\_SIMU for IR and IR+SISR problems. Best results are in bold, while second best results are underlined. Computation times are obtained using a Nvidia Quadro RTX8000 GPU.}
    \label{tab:results_simulated}
\end{table*}
\begin{figure*}[t]
    \centering
    \changebox{
        \includegraphics[width = 0.8\textwidth]{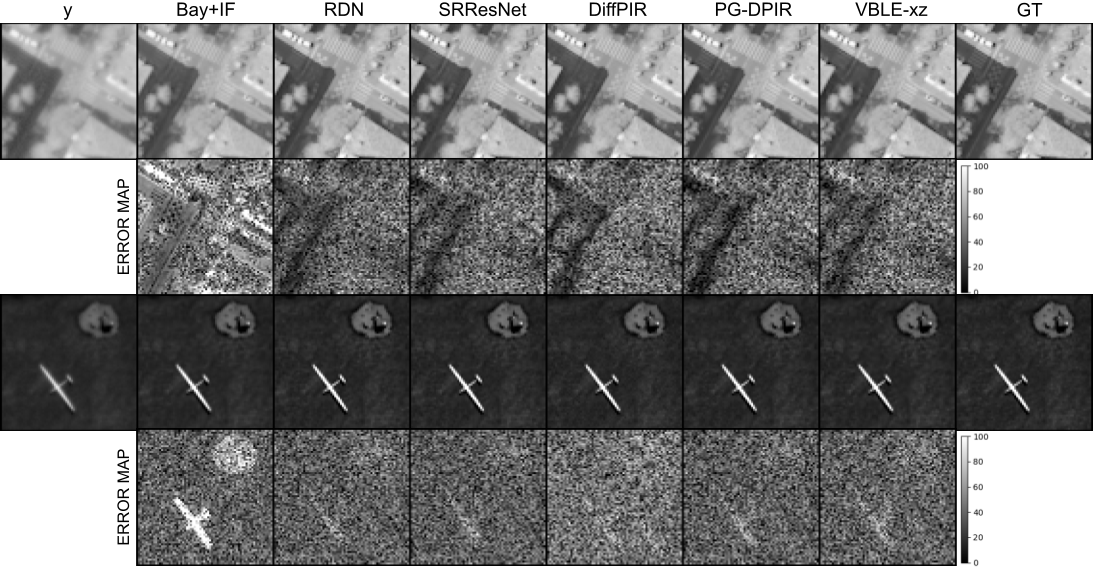}}
    
        \changebox{
        \includegraphics[width = 0.8\textwidth]{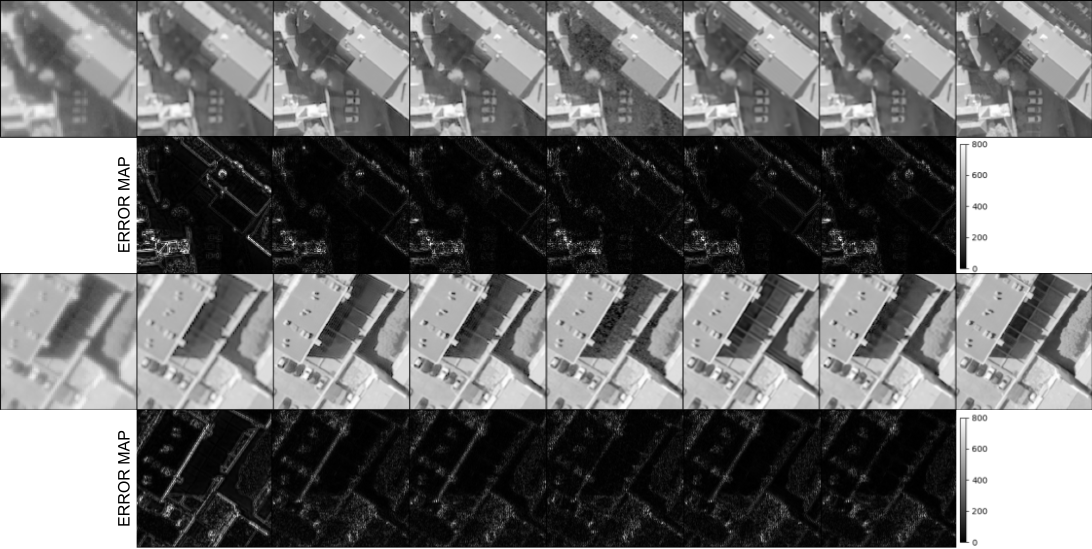}
    }
    \caption{Qualitative results of VBLE-xz and the baselines on TEST30\_SIMU. The first two images correspond to the IR problem, the others correspond to IR+SISR. \change{The error map is the absolute error between the image and GT (Ground Truth).}\copyright CNES 2025}
    \label{fig:visu_simulated}
\end{figure*}

In this section, we present the results of different methods on the TEST30 dataset. In particular, we aim to compare VBLE-xz and PG-DPIR, which are deep regularization (DR) methods, to direct inversion methods RDN and SRResNet, which are mainly chosen in satellite image processing chains. 
Quantitative metrics are presented in \cref{tab:results_simulated} while \cref{fig:visu_simulated} contains visual results.
\Cref{tab:results_simulated} contains results for the image restoration problem, with (IR+SISR) and without super-resolution (IR). 
First, the Bay+IF method is by far outperformed by all deep learning methods. 
VBLE-xz and PG-DPIR outperform DiffPIR and are competitive, or above direct inversion in terms of PSNR and SSIM, while being slightly below them in terms of LPIPS. This is expected because inversion methods are partly trained with a perceptual loss, DISTS, which is similar to LPIPS, whereas DR methods only consider pixelwise likelihood. \change{Interestingly, in the IR task, SRResNet achieves a higher PSNR than RDN, despite RDN being deeper and theoretically expected to perform better. As the test set differs from the trained images, it is likely that SRResNet generalizes better than RDN,  which is plausible given that SRResNet is much shallower than RDN. }
\todel{This can be visually illustrated in Figure 6. Indeed,}

\change{Visual results are available in \cref{fig:visu_simulated}.} Direct inversion methods exhibit sharper results with more high-frequency details, which is favorable to a low LPIPS. However, the added high frequencies are not necessarily correct. Moreover, they show non-desirable oscillating deconvolution artifacts, for example in the 4th image of \cref{fig:visu_simulated}.
VBLE-xz is particularly interesting because its MMSE result is slightly smoother than the estimates of the other methods, but exhibits very few artifacts compared to direct inversion methods, as well as to PG-DPIR, which, for instance, invents some lines in the third image of \cref{fig:visu_simulated}.

The computation time needed to restore one image is also provided in \cref{tab:results_simulated}, except for Bay+IF, which runs on CPU. First, the computation time is much higher when super-resolution is performed, as it is necessary to divide images into patches for all methods except SRResNet. Then, direct inversion are faster than DR methods VBLE-xz and PG-DPIR even if DR method's computation time remains scalable. 
Concerning VBLE-xz, its computation time is better than DiffPIR one, yet remains above direct inversion computation times. 
This is expected as VBLE-xz enables posterior sampling. Yet, it remarkably stays computationally scalable. Besides, recall that this is rarely the case with other posterior sampling methods as depicted in \cref{tab:uncertainty} of the previous section. 

Therefore, PG-DPIR and VBLE-xz provide excellent results, similar to direct inversion methods in terms of performance. This is remarkable as these methods have not been trained to solve a specific inverse problem. Besides, PG-DPIR is very fast for IR, a bit slower than RDN and SRResNet for IR+SISR but remaining computationally efficient. 
VBLE-xz is slightly slower, however, it stays remarkably fast for a posterior sampling method.
%\todel{VBLE-xz is slightly slower, however, it stays remarkably fast for a posterior sampling method.} 
%\change{VBLE-xz is slightly slower, and, even if it is visually interesting, remains slightly below PG-DPIR in terms of metrics. However, we demonstrated  in \cref{sec:posterior_sampling_exp} that VBLE-xz is state-of-the-art among posterior sampling methods, and these results show that its point estimation results are also very competitive even with deterministic baselines.}

\subsection{Results on real satellite images}
\begin{figure*}[t]
        \centering
        \includegraphics[width = 0.8\textwidth]{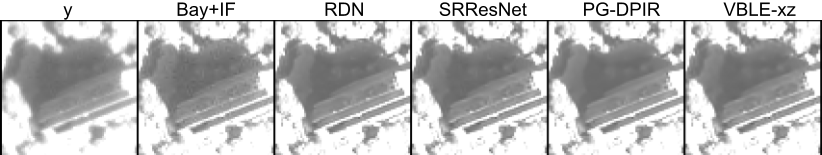}
        \caption{Handling of compression artifacts for each method. Restoration results of several methods on a real DENVER satellite image of TEST30\_REAL, for the IR problem. CI stands for the commercialized image. \copyright CNES 2025}
        \label{fig:visu_real_70}
        \centering
        \includegraphics[width = 1\textwidth]{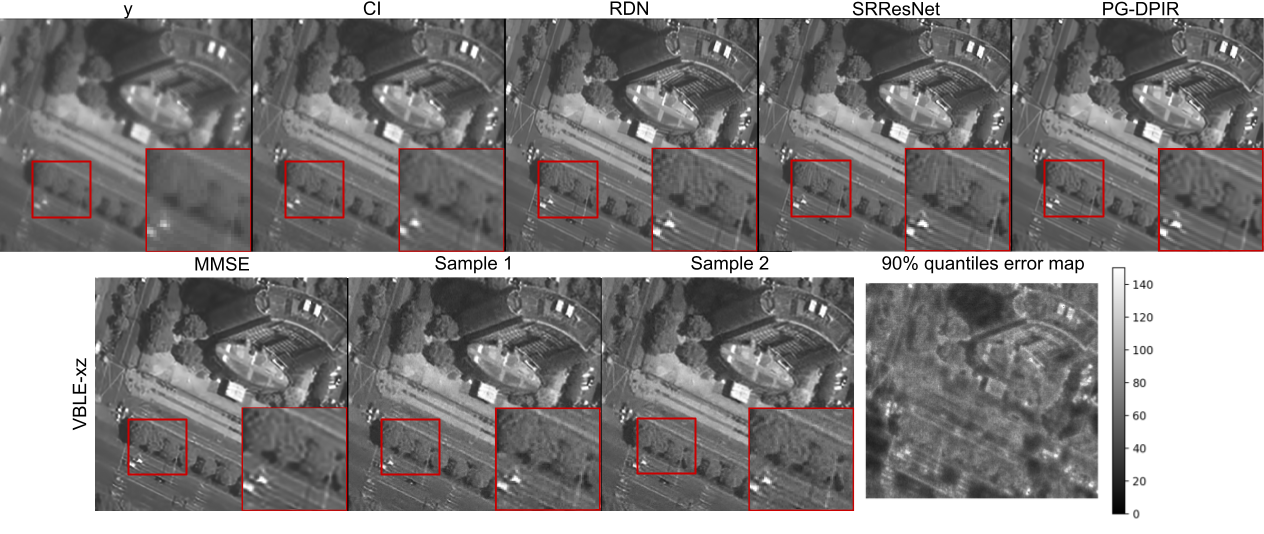}
        \caption{Results for several methods on a real DENVER satellite image of TEST30\_REAL, for IR+SISR problem. CI stands for commercialized image.\copyright CNES 2025}
        \label{fig:visu_real_35}
    \end{figure*}
In this section, we present the results of the different methods on TEST30\_REAL. As we do not have access to any ground truth, the analysis will be essentially based on visual results. Additionally, we provide BRISQUE scores in \cref{tab:brisque}, though this metric exhibits high variance and might not be fully adapted to satellite images. 
CI refers to the currently Commercialized (Pléiade) Image. Direct inversion methods exhibit the best BRISQUE scores for the IR+SISR problem, which can be explained by their sharp results involving "typical" high frequencies that are not necessarily accurate. However, in IR, VBLE-xz is competitive with the direct inversion baselines. 

\begin{table}[h]
    \label{tab:brisque}
    \centering
    \setlength{\tabcolsep}{4pt}
    \scriptsize{
        \begin{tabular}{clcccccc}
            \hline
            Pb. &  & $y$ & \textbf{CI} & \textbf{RDN} & \textbf{SRResNet} & \textbf{DPIR} & \textbf{VBLE-xz} \\ \hline
            \multirow{4}{*}{\rotatebox{90}{IR}} & ATAC. & 41.20 & 51.09 & \textbf{24.85} & 50.11 & 46.96 & \underline{32.27} \\
             & DENV. & 27.48 & 27.88 & 30.34 & \underline{23.89} & 24.74 & \textbf{20.81} \\
             & SHAN. & 32.43 & 31.93 & \underline{27.21} & 27.90 & 31.40 & \textbf{26.19} \\
             \cdashline{2-8}
             & TOTAL & 33.70 & 36.97 & \textbf{24.13} & 33.97 & 34.37 & \underline{26.42} \\ \hline
            \multirow{4}{*}{\rotatebox{90}{IR+SISR}
            } & ATAC. & x & 56.30 & \textbf{19.13} & \underline{30.51} & 54.57 & 36.39 \\
             & DENV. & x & 40.15 & \textbf{5.42} & \underline{14.39} & 36.47 & 40.16 \\
             & SHAN. & x & 39.31 & \textbf{5.79} & \underline{12.64} & 42.64 & 40.15 \\
            \cdashline{2-8}
             & TOTAL & x & 45.25 & \textbf{10.11} & \underline{19.18} & 44.56 & 38.90 \\ \hline
            \end{tabular}
            \caption{BRISQUE $\downarrow$ scores on TEST30\_REAL. ATAC., DENV. and SHAN. stand resp. for Atacama, Denver and Shanghai landscapes. CI stands for the commercialized image.}
    }
    % \vspace{-4em}
    \end{table}

Visual results are given in \cref{fig:visu_real_70} and in  \cref{fig:visu_real_35}. In \cref{fig:visu_real_70}, the measure contains some compression artifacts which are amplified by the restoration performed in the commercialized image (CI). All deep learning methods perform better than CI for removing these artifacts. However, for direct inversion methods RDN and SRResNet, some of the artifacts still slightly remain, while DR methods manage to remove them entirely. Indeed, direct inversion methods yield sharper results but deconvolve details that do not lie in the true image. 

\looseness=-1
\Cref{fig:visu_real_35} shows the results of image restoration with super-resolution for all methods. For VBLE-xz, the MMSE estimate, as well as two posterior samples and the predicted 90\% perpixel quantiles are provided. The zoomed commercialized image is very blurry, direct inversion methods yield sharp results, but with some artifacts, while DR methods yield smoother results. PG-DPIR provides excellent results, however, it sometimes yields lines that do not exist. VBLE-xz exhibits very consistent results, its MMSE estimate is smoother than images obtained by direct inversion, but the samples are sharper and contain valuable information, for instance concerning the shapes of the linears. 
Besides, the 90\% quantile map provides a meaningful way to localize unsure areas. 

\subsection{Robustness evaluation}

In this section, we evaluate the robustness of the different methods to resolution and \todel{model changes}\change{modeling errors} using TEST30. 
% \subsubsection{Domain robustness}
% \change{
% First, it is important to emphasize that DR and direct inversion methods differ significantly in terms of domain adaptation. While simulating realistic target images is feasible when very high-resolution airborne data are available, generating realistic degraded images is much more challenging. This difficulty arises because not all physical phenomena can be accurately modeled, and the blur is acquisition-dependent, varying with satellite orientation. As a result, a domain gap exists when training direct inversion methods with pairs of simulated target and degraded images. DR methods, by contrast, do not suffer from this domain gap since they are trained solely on target images. %However, they rely on the satellite forward model during restoration, which is not exact, making modeling errors likely in practice.
% }

\subsubsection{Robustness to resolution change}

\begin{figure}[h]
    \centering
    \includegraphics[width = 0.5\textwidth]{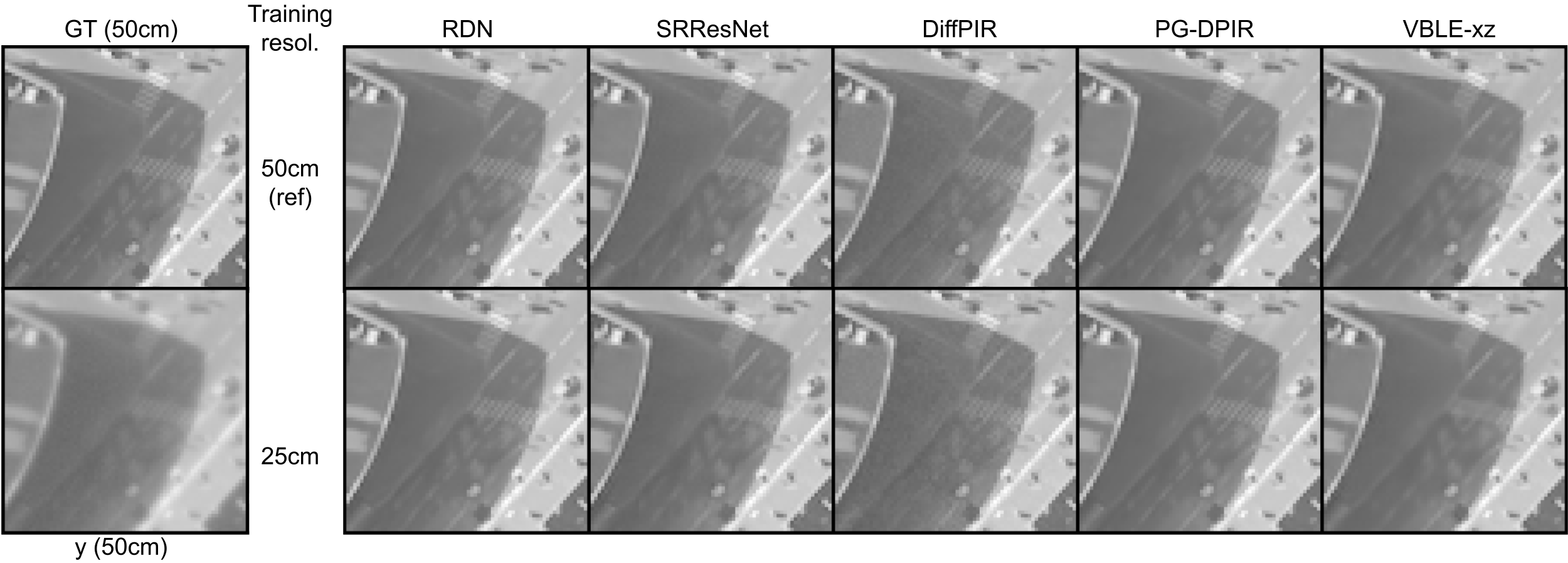}
    \caption{Restored image from TEST30 by several methods for the IR problem, using a network trained at the right resolution (first row), and a network trained at the wrong resolution (second row). \copyright CNES 2025}
    \label{fig:visu_resol_50}
\end{figure}

Direct inversion models are problem-specific, therefore they need to be retrained or fine-tuned for each resolution and each inverse problem. For DR methods VBLE-xz and PG-DPIR, the models can be used to restore several inverse problems, yet, it should \textit{a priori} be retrained for each resolution. 
In the following experiments, we evaluate the robustness to resolution change for IR and IR+SISR problems. For IR, we use the DR models trained on 25cm target images and evaluate them on 50cm images. For direct inversion methods, RDN and SRResNet are trained to deblur 25cm images and are evaluated on 50cm images. For IR+SISR, we use the DR models trained at 50cm and evaluate them on 25cm images. 
    \begin{table*}[t]
        \centering
        \scriptsize
        \begin{tabular}{l|c|ccc|c|ccc}
            \hline
             & \multirow{2}{*}{\begin{tabular}[c]{@{}c@{}}Training\\ Resol.\end{tabular}} & \multicolumn{3}{c|}{\textbf{IR}} & \multirow{2}{*}{\begin{tabular}[c]{@{}c@{}}Training\\ Resol.\end{tabular}} & \multicolumn{3}{c}{\textbf{IR+SISR}} \\
             &  & PSNR $\uparrow$ & SSIM $\uparrow$ & LPIPS $\downarrow$ &  & PSNR $\uparrow$ & SSIM $\uparrow$ & LPIPS $\downarrow$ \\ \hline
            \multirow{2}{*}{RDN} & 50 (ref) & 48.08 & 0.9951 & 0.0145 & 25 (ref) & x & x & x \\
             & 25 & 48.00 & 0.9951 & 0.0144 & 50 & x & x & x \\ \hdashline
            \multirow{2}{*}{SRResNet} & 50 (ref) & 48.44 & 0.9949 & 0.0157 & 25 (ref) & x & x & x \\
             & 25 & 48.16 & 0.9948 & 0.0165 & 50 & x & x & x \\ \hdashline
             \multirow{2}{*}{DiffPIR} & 50 (ref) & 45.48 & 0.9893 & 0.0464 & 25 (ref) & 36.26 & 0.9345 & 0.1930 \\
             & 25 & 45.42 & 0.9892 & 0.0488 & 50 & 35.98 & 0.9327 & 0.1913 \\ 
             \hdashline
            \multirow{2}{*}{PG-DPIR} & 50 (ref) & 48.66 & 0.9952 & 0.0138 & 25 (ref) & 37.18 & 0.9513 & 0.1658 \\
             & 25 & 48.78 & 0.9953 & 0.0142 & 50 & 37.31 & 0.9542 & 0.1527 \\ \hline
            \multirow{2}{*}{VBLE-xz} & 50 (ref) & 48.31 & 0.9948 & 0.0198 & 25 (ref) & 36.65 & 0.9486 & 0.1708 \\
             & 25 & 47.99 & 0.9945 & 0.0199 & 50 & 36.56 & 0.9479 & 0.1733 \\ \hline
        \end{tabular}
        \caption{Quantitative results for resolution robustness experiments on TEST30\_SIMU. For each method and each problem, the first line corresponds to the results using the network trained at the right resolution, while the second line denotes the results using the network trained at the wrong resolution.}
        \label{tab:resolution}
    \end{table*}
Quantitative results are provided in \cref{tab:resolution}. Surprisingly, for all methods, the performance is not significantly affected by the resolution change. This could be explained by the fact that some invariance to resolution exists in several satellite landscapes, for instance in textured areas. However, it is not the case for some small or high-frequency structures, such as cars or pedestrian crossings. In \cref{fig:visu_resol_50}, which provides the restoration of a pedestrian crossing for each method at the two resolutions, the crossing is restored worse with the model trained at the wrong resolution. This is particularly visible for PG-DPIR as it restores particularly well the crossing at the right resolution.

% \subsubsection{Robustness to model changes}
\subsubsection{Robustness to modeling errors}
\begin{figure}[h]
    \begin{minipage}[b]{0.49\linewidth}
        \centering
        \includegraphics[width=\textwidth]{"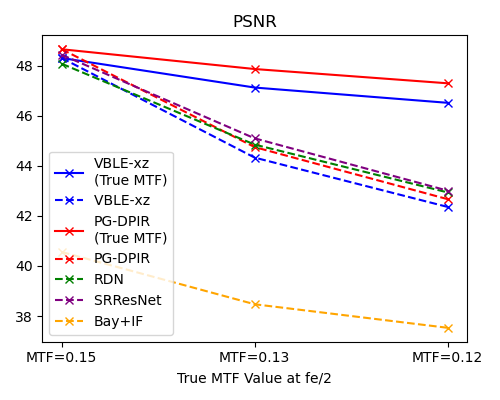"}      
       \end{minipage}
       \begin{minipage}[b]{0.49\linewidth}
        \centering
        \includegraphics[width=\textwidth]{"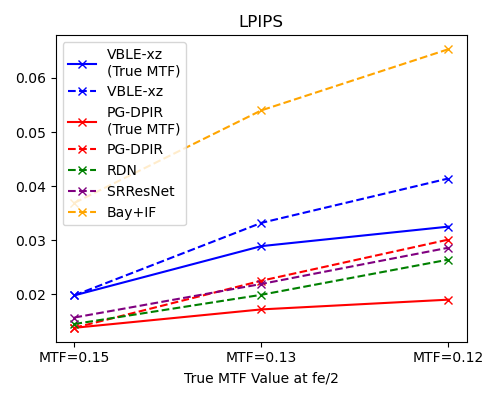"}      
       \end{minipage}

       \caption{PSNR and LPIPS of several methods with changes in the forward model. Dashed lines denote the results without modeling the change. For DR methods PG-DPIR and VBLE-xz, continuous lines denote the results when the change is modeled in the forward model during restoration. MTF=0.15 at $f_e/2$ is the original value of the MTF in the experiments.}
       \label{fig:ftm_change}
       \vspace{-1.7em}
\end{figure}

\begin{figure}[h]
    \centering
    \changebox{
    \includegraphics[width = 0.45\textwidth]{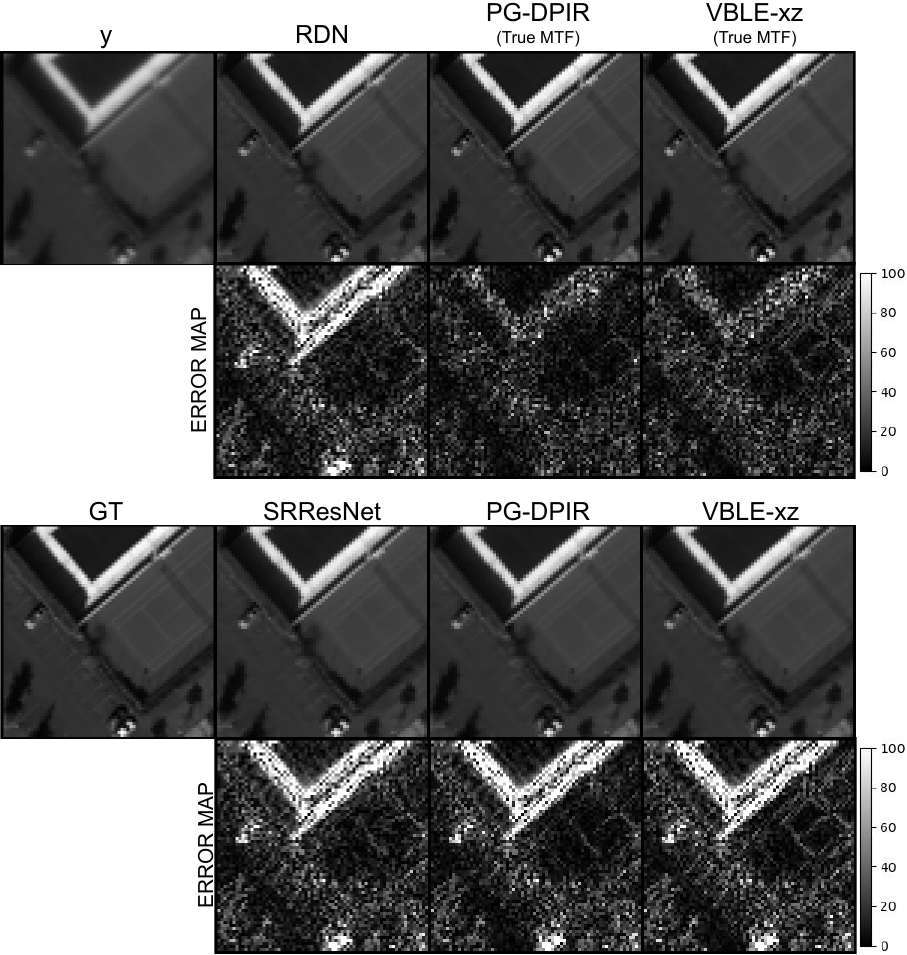}
    }
    \caption{Restored image from TEST30 by several methods for the IR problem with changes in the degradation model (MTF=0.12 at $f_e/2$). This forward model change is not modeled, except for DR methods when \emph{True MTF} is mentioned. 
    \copyright CNES 2025}
    \label{fig:visu_changes}
\end{figure}

Here, we evaluate the robustness of the different methods to \todel{changes in the forward model} modeling errors. 
\change{
% First, it is important to emphasize that the performance gap induced by the training on simulated data when restoring real-world image significantly differs for DR and direct inversion methods. 
Note that testing on real images a model trained only on simulated images is a disadvantageous scenario for performance assessment. However, it should be emphasized that it affects differently DR and direct inversion 
methods.  Indeed, simulating realistic target images is quite feasible when very high-resolution airborne data are available. On the opposite, generating realistic degraded images is much more challenging. This difficulty arises because not all the physical phenomena can be accurately modeled. For instance, the blur is acquisition-dependent and varies with the satellite orientation. As a result, a domain gap exists when training direct inversion methods with pairs of simulated target and degraded images. DR methods, by contrast, do not seem to suffer from this domain gap since they are trained solely on target images. However, note that DR methods are impacted by the modeling errors regarding the satellite forward model during restoration step. As quantifying the performance gap occurring when restoring real data  is hardly feasible as no ground truth data are available, we choose, 
}
%We choose, 
for simplicity, to induce slight changes in the PSF, parameterized by the Modulated Transfer Function (MTF), that is the Fourier transform of the PSF, evaluated at $f_e/2$. The Pléiades MTF at $f_e/2$ is $0.15$. We compute image restoration results for MTF values of $0.13$ and $0.12$ at $f_e/2$. These values are representative of the evolution of the instrument PSF in orbit. 
As direct inversion methods are trained for a specific PSF with MTF $=0.15$, they should not be very robust to these changes. For DR methods, if the change of PSF is taken into account in the forward model during the restoration, the performance should not be affected. A visual illustration is given in \cref{fig:visu_changes}, while quantitative results are provided in \cref{fig:ftm_change}.

Dashed lines in the figure represent the performance of the methods when the wrong PSF, with MTF$=0.15$, is used for image restoration. Continuous lines for DR methods represent the performance when the PSF change is integrated into the forward model. When the wrong PSF is used for DR methods, the performance decay is at the same rate as for RDN and SRResNet methods. Yet, when the PSF change is integrated into the forward model, the performance is not affected, showing that DR methods can be used with different PSFs without any retraining, as long as the forward model is correctly specified. 
This can be crucial in practice as the PSF of a satellite evolves with time, mainly due to focus changes. Conversely, direct inversion networks should be retrained each time the PSF is significantly modified.

Therefore, PG-DPIR and VBLE-xz are robust to changes in the resolution and in the forward model. Additionally, they also work very well on real data, as shown in the previous section. These data contain new types of landscapes that have not been seen during training, \text{i.e.} clouds and Atacama desert, and additional model errors induced by compression and satellite off-pointing. This shows that the methods are robust to a wide range of problems and can be used in practice in real satellite processing chains.

\subsection{Ablation study}

\begin{figure}[h]
    \centering
    \includegraphics[width = 0.4\textwidth]{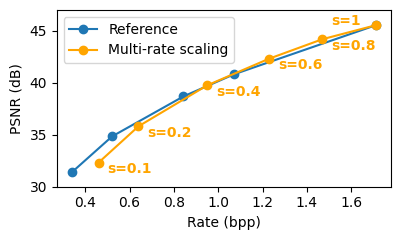}
    \caption{Rate-distortion curve computed on TEST30. Blue curve: each point corresponds to a CAE trained at a specific bitrate. Orange curve: One CAE trained at a high bitrate, with different multirate scaling factors.}
    \label{fig:rate_distortion}
\end{figure}

In this section, we demonstrate the usefulness of the additional scaling factor $s$, which enables the modulation of the CAE bitrate, and thus of the regularization strength when solving different inverse problems.  
\Cref{fig:rate_distortion} shows two rate-distortion curves obtained on TEST30, illustrating the trade-off between compression efficiency and reconstruction error for a given compressor. The blue curve corresponds to the rate-distortion trade-offs of several single-rate CAEs, each trained with a different $\alpha$, as defined in \cref{eq:cae_loss}. The orange curve represents the trade-offs obtained using the CAE trained for the highest bitrate, while varying the scaling factor $s$ to modulate the rate, following the approach described in \cite{Arezki2024} and applied in VBLE-xz. For $s \in [0.2, 1]$, the two curves are nearly identical, validating the effectiveness of this multirate approach for satellite data.

\begin{table}[h]
\footnotesize
\begin{tabular}{lcccc}
\hline
 & PSNR & SSIM & LPIPS & ICP 90\% \\ \hline
CAE $\alpha=0.0932,s=1$ & 37.35 & 0.9490 & 0.1698 & 0.87 \\
CAE $\alpha=0.36, s=0.4$ & 37.31 & 0.9484 & 0.1691 & 0.88 \\ \hline
\end{tabular}
\caption{Results of IR+SISR on TEST14. $\alpha$ is the rate-distortion parameter of \cref{eq:cae_loss}, a high $\alpha$ corresponds to a high bitrate. $s$ is the multirate scaling factor.}
\label{tab:scale_vs_rate}
\end{table}

Additionally, some image restoration results on the IR+SISR problem are provided in \cref{tab:scale_vs_rate}, comparing the use of a CAE trained at an adapted low bitrate with $s=1$ and a CAE trained at a high bitrate with $s=0.4$. Both networks yield very similar performance. This demonstrates the relevance of incorporating the scaling factor $s$ in VBLE-xz. Indeed, it enables control over the regularization through a single additional hyperparameter, rather than requiring the training of another CAE with an appropriate rate-distortion trade-off.

\section{Conclusion}
\label{sec:conclusion}

In this paper, we proposed the VBLE-xz method for satellite image restoration. VBLE-xz estimates the inverse problem posterior using variational inference by capturing the uncertainty jointly in the latent and the image space of a variational CAE. 
We also proposed to use a scaling factor to modulate the CAE bitrate, a remarkably simple way to adapt the regularization to a given inverse problem difficulty.
 
\change{
While deep regularization (DR) methods possess various advantages compared to direct inversion methods - in particular restoring images from different sources with a single network, and avoiding the use of simulated degraded images for training -, our experiments demonstrate that the DR methods VBLE-xz and PG-DPIR also present advantages in terms of restoration performance. In particular, they provide consistent performance at a reasonable computational cost and introduce fewer artifacts than direct inversion. When uncertainty quantification (UQ) is required, VBLE-xz offers the best trade-off between image quality, UQ accuracy, and computation time. In contrast, when UQ is not needed, PG-DPIR achieves slightly better performance, yet with more pronounced artifacts.
}
\todel{
Experiments were conducted either on realistic simulated or on real satellite images. 
These experiments demonstrated the interest of VBLE-xz as a scalable posterior sampling method for satellite image restoration.
VBLE-xz is indeed by several orders of magnitude faster than the posterior sampling baselines, while yielding similar or better performance and uncertainty quantification.
Furthermore, the experiments demonstrated the interest of the deep regularization methods VBLE-xz and PG-DPIR over direct inversion methods which are classically used in the satellite processing chain. Indeed, VBLE-xz and PG-DPIR yield consistent performance, acceptable computation cost and exhibit very few artifacts compared to them, while enabling the restoration of images from several sources with the same network, unlike them. 
Hence, they represent very interesting alternatives to direct inversion methods for real image processing chains, particularly VBLE-xz when uncertainty quantification is required. }

Future works will be dedicated to improving and extending the method, for instance, by introducing correlations in the image space uncertainty model. 
Another interesting perspective will be to demonstrate the interest of the generated uncertainties for real remote sensing use cases such as change detection or land-cover mapping.

\section*{Acknowledgments}
This work was partly supported by CNES under project name DEEPREG, and ANITI under grant agreement ANR-19-PI3A-0004.

% \bibliographystyle{abbrv}
% \bibliography{biblio}
\FloatBarrier
{
    \small
    \bibliographystyle{abbrv}
    \bibliography{biblio}
}

\end{document}